\documentclass{article}

\usepackage[final]{neurips_2022}
\usepackage{lipsum}
\usepackage{listings}
\usepackage{caption}
\usepackage{algpseudocode}
\usepackage{algorithm}

\usepackage{subfigure}
\usepackage{microtype}
\usepackage{graphicx}
\usepackage{amsmath}
\usepackage{amssymb}
\usepackage{mathtools}
\usepackage{amsthm}
\usepackage{todonotes}

\usepackage[utf8]{inputenc} 
\usepackage[T1]{fontenc}    
\usepackage{hyperref}       
\usepackage{url}            
\usepackage{booktabs}       
\usepackage{amsfonts}       
\usepackage{nicefrac}       
\usepackage{microtype}      
\usepackage{xcolor}         

\title{Repeated Augmented Rehearsal: A Simple but Strong Baseline for Online Continual Learning}

\author{%
  Yaqian Zhang\\
  University of Waikato\\
  New Zealand \\
  \texttt{yaqianz@waikato.ac.nz} \\
  \And
    Bernhard Pfahringer\\
    University of Waikato \\
  New Zealand \\
  \texttt{bernhard@waikato.ac.nz} \\
  \And
  Eibe Frank \\
    University of Waikato \\
  New Zealand \\
  \texttt{eibe@waikato.ac.nz} \\
  \And
  Albert Bifet \\
   University of Waikato \\
  New Zealand \\
  LTCI, Télécom Paris, France\\
  \texttt{abifet@waikato.ac.nz} \\
  \And
  Nick Jin Sean Lim \\
    University of Waikato \\
  New Zealand \\
  \texttt{nick.lim@waikato.ac.nz} \\
  \And
  Yunzhe Jia\\
    University of Waikato \\
  New Zealand \\
  \texttt{alvin.jia@waikato.ac.nz} \\
}

\begin{document}

\maketitle

\begin{abstract}


Online continual learning (OCL) aims to train neural networks incrementally from a non-stationary data stream with a single pass through data. Rehearsal-based methods attempt to approximate the observed input distributions over time with a small memory and revisit them later to avoid forgetting. Despite their strong empirical performance, rehearsal methods still suffer from a poor approximation of past data’s loss landscape with memory samples. This paper revisits the rehearsal dynamics in online settings. We provide theoretical insights on the inherent memory overfitting risk from the viewpoint of biased and dynamic empirical risk minimization, and examine the merits and limits of repeated rehearsal.
Inspired by our analysis, a simple and intuitive baseline, repeated augmented rehearsal (RAR), is designed to address the underfitting-overfitting dilemma of online rehearsal. Surprisingly, across four rather different OCL benchmarks,
this simple baseline outperforms vanilla rehearsal by  9\%-17\% and also significantly improves the state-of-the-art rehearsal-based methods MIR, ASER, and SCR. We also demonstrate that RAR successfully achieves an accurate approximation of the loss landscape of past data and high-loss ridge aversion in its learning trajectory. Extensive ablation studies are conducted to study the interplay between repeated and augmented rehearsal, and reinforcement learning (RL) is applied to dynamically adjust the hyperparameters of RAR to balance the stability-plasticity trade-off online. Code is available at \href{https://github.com/YaqianZhang/RepeatedAugmentedRehearsal}{https://github.com/YaqianZhang/RepeatedAugmentedRehearsal}.

\end{abstract}

\section{Introduction}
\label{intro}
Despite its recent success, deep learning largely relies on the assumption of independent and identically distributed (i.i.d.) data that can be repeatedly revisited during training.  
 Non-i.i.d settings are challenging for neural networks due to catastrophic forgetting: previously learned knowledge can easily be overwritten when training on new data because this data may follow a different distribution~\citep{li2017learning,rebuffi2017icarl,delange2021continual}.
Online continual learning (OCL or Online CL) studies how to enable deep learning in an online manner from a non-stationary data stream. As the data stream can be vast or even infinite, it is infeasible to store and shuffle the dataset for multiple epochs of training. Therefore, a fundamental assumption is that the data stream can only be accessed one batch at a time and training is performed with a \textit{single pass} over the data~\citep{aljundi2019online}.

Experience replay (ER), also known as rehearsal~\citep{chaudhry2019tiny,delange2021continual}, is a key idea in OCL. It stores a subset of previously seen data $\mathcal{D}$ in a fixed-size memory $\mathcal{M}$ and revisits the memorized samples during training to mitigate forgetting of previous tasks. To update the model, a batch sampled from the memory is combined with the incoming batch from the stream to compute the gradient~\citep{chaudhry2019tiny}. 
Different variants of ER have been developed to improve memory management policies and representation learning, achieving state-of-the-art performance in a number of standard OCL benchmarks ~\citep{aljundi2019online,mai2021supervised,shim2021online}.

However, whether rehearsal is appropriate for continual learning, considering the risk of overfitting the memory when using data from the memory to directly contribute to the gradient computation, has been debated vigorously~\citep{lopez2017gradient,chaudhry2019tiny,verwimp2021rehearsal}. The potential for overfitting has motivated the development of constraint-based replay methods~\citep{lopez2017gradient}, e.g., GEM and A-GEM, which use memory samples solely to {\em constrain} the gradient direction. However, there is empirical evidence suggesting that rehearsal-based methods consistently outperform methods that do not train directly on the memory~\citep{chaudhry2019tiny}. This indicates rehearsal on the memory does not necessarily prevent effective generalization, possibly due to the implicit regularization effect of incoming data~\citep{chaudhry2019tiny}. Nevertheless, recent work analyzing the loss landscape when applying rehearsal to offline continual learning finds that memory samples indeed provide
a poor approximation of the loss landscape of past tasks, especially near a high-loss ridge. As a result, ``instead of ending up {\em near} the high-loss ridge in perspective of the rehearsal memory, the solution in reality
resides {\em on} the high-loss ridge for the training data''~\citep{verwimp2021rehearsal}. This latest finding poses the question of how to better approximate the loss surface of past data $\mathcal{L}(\mathcal{D};\theta)$ with memory samples' loss $\mathcal{L}(\mathcal{M};\theta)$.

To better approximate past data's loss surface, previous work studies which samples should be memorized for rehearsal. Instead, we examine the optimization process during rehearsal and study how to effectively perform rehearsal with the memorized samples. Focused on \textit{online} CL, our study extends the previous understanding of rehearsal along two directions. First, we provide theoretical considerations that reveal two insights regarding the extent of overfitting to memory: it is a) related to the inherent attributes of the OCL problem concerned and b) varies across the different stages of continual learning. Second, we highlight the limits of applying rehearsal with multiple iterations---a trick used to maximally utilize the incoming batch~\citep{aljundi2019online}---and identify an underfitting-overfitting dilemma for online rehearsal.


Based on our analysis, we design a simple baseline 
to deal with the underfitting-overfitting dilemma in online CL problems, dubbed repeated augmented rehearsal (RAR), that can be easily integrated into existing rehearsal-based methods. Surprisingly, this simple baseline leads to a large performance boost for ER, as well as state-of-the-art ER-based approaches, across four OCL benchmarks. More importantly, the loss landscape analysis shows that RAR can help memory samples  reliably approximate the  distribution of past data and successfully avoids the high-loss ridge of past tasks. To better understand the behavior of RAR, we further investigate the interplay between repeated rehearsal and augmented rehearsal via an ablation study. We also propose a reinforcement learning-based method to dynamically adjust the hyperparameters of RAR and balance the stability-plasticity trade-off in an online manner.




\section{Related Work}
\label{sec:related_work_OCL}
\textbf{Online continual learning}: We consider the online continual learning setting with a non-stationary (potentially infinite) stream of data $\mathcal{D}_t$: at each time step $t$, the continual learning agent receives an
incoming batch of data samples
$\mathcal{B}_t=\{\textbf{x}_i,y_i\}_{i=1,..,|\mathcal{B}_t|}$
that are drawn from the current data distribution $\mathbb{P}(\mathcal{D}_t)$. The period of time where the data distribution stays the same is often called a \textit{task} or \textit{experience} in the continual learning literature. An abrupt change in the data distribution occurs when the task changes. 
The standard objective during training is to minimize the empirical risk on all the data seen so far:
\begin{equation}
\label{eq:loss_cl}
   \min_\theta \mathcal{R}(\theta)=\min_\theta\frac{1}{\sum_t|\mathcal{B}_t|}\sum_{t} \sum_{\textbf{x},y\in \mathcal{B}_t} \mathcal{L}\left(f_\theta(\textbf{x}),y \right),
\end{equation}
with loss function $\mathcal{L}$, the CL network function $f$, and its associated parameters $\theta$. 

\textit{Metrics}:
A common metric is the end accuracy after training on $T$ tasks, defined as 
$\textstyle{A_T = \frac{1}{T}\sum^{j=T}_{j=1}a_{T,j}}$,
where $a_{i,j}$ denotes the model's accuracy on the held-out test set of task $j$ after training on task $i$. Other metrics are ``forgetting''~\citep{chaudhry2018efficient}, which is defined as $\textstyle{F_T=-\frac{1}{T-1} \sum_{i=1}^{T-1}\left(a_{T, i}-\max _{l \in 1 \ldots T-1} a_{l, i}\right)}$ and the related metric ``backward transfer''~\citep{lopez2017gradient}: 
$\textstyle{{B_T}=\frac{1}{T-1} \sum_{i=1}^{T-1} a_{T, i}-a_{i, i}}$.

\textit{Online setting}:
A key difference between online and offline CL is that the latter assumes full access to the whole training data for the task that is currently being processed. Therefore, it allows training on each single task with multiple epochs (e.g., 70-200 epochs) \citep{rebuffi2017icarl,wu2019large,cha2021co2l}. The online CL setting is more challenging because the agent can only access the current batch of incoming data and performs training with a single pass through the data.  

\textbf{Experience Replay (Rehearsal)}: Chaudhry et al. (\citeyear{chaudhry2019tiny}) propose experience replay (ER), which performs joint training on memory samples and incoming samples. A simple but strong baseline approach to sampling in ER is reservoir sampling~\cite{vitter1985random}. Aljundi et al. (\citeyear{aljundi2019online}) propose Maximally Interfered Retrieval (MIR), which retrieves the samples that will be most negatively impacted by the foreseen parameter
updates. Shim et al. (\citeyear{shim2021online}) propose ASER, which selects samples to best preserve existing memory-based class boundaries. In terms of model training, Mai et al. (\citeyear{mai2021supervised}) propose to replace the cross-entropy loss with the supervised contrastive loss to learn a better representation. We consider all these variants of ER in our experiments.

\textit{Augmentation}: In the standard i.i.d. setting, data augmentation is a widely used method to improve deep learning~\citep{cubuk2020randaugment}. In the offline CL setting,~\cite{mai2021supervised} use augmentation  to construct a supervised contrastive loss and \cite{bang2021rainbow} employ it together with an uncertainty-based memory management strategy. However, these papers apply augmentation together with other advanced techniques, and there is no ablation study on the effect of augmentation per se. Thus, it is unclear whether augmentation itself is beneficial to rehearsal or not, especially in the online setting.

\textit{Repeated Rehearsal (Multiple Iterations)}: Vanilla online continual learning employs a single gradient update given an incoming batch of data. To maximally utilize the current incoming batch, \cite{aljundi2019online} propose to perform multiple gradient updates instead. Their experiment with CIFAR10 shows using MIR with five iterations leads to a 1.7\% improvement in accuracy. In this paper, we systematically analyze the effect of multiple iterations on online rehearsal and provide theoretical and empirical insights on when this trick may improve or harm performance.

\textbf{Hyperparameter Tuning for OCL}: Hyperparameter tuning is a particular challenge in OCL due to the lack of a dedicated validation set and the constraint of a single pass through the data. \citet{chaudhry2018efficient} and \citet{mai2022online} employ a hyperparameter tuning protocol that uses an external validation data stream with a small number of tasks. Offline hyperparameter tuning is applied to this validation data with multiple passes to identify optimal values, which are then used for the actual online continual learning tasks. A limitation of this method is that it relies on external validation data. 

\section{Revisiting Online Rehearsal: Is Repeated Rehearsal a Good Idea?  }
\label{sec:lugo}

We revisit rehearsal from two directions. First, while previous work demonstrates its strong empirical performance~\citep{chaudhry2019tiny} and provides conceptual analysis for \textit{offline} CL~\citep{verwimp2021rehearsal}, we  focus on the \textit{online} setting and provide  theoretical insights through the lens of empirical risk minimization (ERM). Second, we examine the dynamics of employing rehearsal with multiple iterations. This has been proposed as a trick for online CL to maximally utilize the incoming batch already ~\citep{mai2022online,aljundi2019online}; we investigate whether it is always better than rehearsal with a single iteration.
\label{sec:algo}

 \subsection{Empirical Risk Minimization in Online Rehearsal: a Biased and Dynamic Objective}
 \label{sec:erm}
For rehearsal in OCL, at each iteration $t$, a batch of data $\mathcal{B}_t$ is obtained from the incoming task, where $\mathcal{B}_t\sim \mathcal{D}_\mathcal{T}$ and $\mathcal{B}_t=\{\textbf{x}_i,y_i\}_{i=1,...|\mathcal{B}_t|}$, and a batch $\mathcal{B}^\mathcal{M}_t$ is sampled from memory, where $\mathcal{B}^\mathcal{M}_t\sim \mathcal{D}^t_\mathcal{M}$ and $\mathcal{B}^\mathcal{M}_t=\{\textbf{x}_i,y_i\}_{i=1,...|\mathcal{B}^\mathcal{M}_t|}$. 
The gradient-based update rule of ER is:
  \begin{equation}
 \label{eq:grad_er}
   \theta_{t+1}=\theta_t-
   \frac{\eta}{\left|\mathcal{B}_{t}\right|} 
   \sum_{\textbf{x},y \in  \mathcal{B}_t} \nabla \mathcal{L}\left(f_\theta(\textbf{x}),y\right) -\frac{\eta}{\left|\mathcal{B}^\mathcal{M}_{t}\right|} \sum_{\textbf{x},y \in \mathcal{B}^\mathcal{M}_{t}} \nabla \mathcal{L}\left(f_\theta(\textbf{x}),y\right).  
 \end{equation}
 Given this update rule, we would like to establish the corresponding objective function, but this is not straightforward to derive because the memory is immediately updated after each incoming batch, which means the memory data samples $\mathcal{D}^t_{\mathcal{M}}$ changes all the time. For a widely used memory management policy, where the memory is updated using reservoir sampling~\citep{vitter1985random,chaudhry2019tiny}, we prove in the appendix that the empirical risk  for online ER follows Proposition 1.
 
 \noindent\textbf{Proposition 1 (ERM for online rehearsal)}: Assume an incoming task stream $\mathcal{D}_\mathcal{T}$ and an initial memory set  $\mathcal{D}^0_\mathcal{M}$ with different data distribution $\mathbb{P}(\mathcal{D}_\mathcal{T})\neq \mathbb{P}(\mathcal{D}^0_\mathcal{M})$. Assume further that the memory is updated at the end of each iteration using reservoir sampling. 
 Then, Eq~\ref{eq:grad_er} implements unbiased stochastic gradient descent for the following loss function:

 \begin{equation}
 \label{eq:loss_er}
  \min_\theta\mathcal{R}_t(\theta)= \min_\theta\sum_{\textbf{x},y\in \mathcal{D}_\mathcal{T}} \mathcal{L}(f_\theta(\textbf{x}),y)+\beta_t \lambda\sum_{\textbf{x},y \in \mathcal{D}^0_\mathcal{M}}\mathcal{L}(f_\theta(\textbf{x}),y), 
 \end{equation} 
where $\lambda := \frac{|\mathcal{D}_\mathcal{T}|}{|\mathcal{D}^0_\mathcal{M}|}$ and $|\mathcal{D}^0_\mathcal{M}|$ and $|\mathcal{D}_\mathcal{T}|$ are the memory size and incoming task data size respectively; $\beta_t :={1}/(1+\frac{2N^t_{cur}}{N^\mathcal{T}_{past}})$, and $N^t_{cur} = \sum_{i=1}^{i=t}|\mathcal{B}_i|$ denotes the number of samples of the current task that have been seen so far and  $N^{\mathcal{T}}_{past}=\sum_{j=1}^{j=\mathcal{T}}|\mathcal{D}_j|$ denotes the number of samples pertaining to the tasks so far, excluding the current task. We immediately have $\beta_t\in (0,1]$.


The proposition reveals several interesting properties:
\begin{itemize}
    \item \textbf{Bias}: compared with the true objective in continual learning in Eq~\ref{eq:loss_cl},  the loss of online ER in Eq~\ref{eq:loss_er} is actually a biased approximation of the former, as it puts a different weight ($\beta_t\lambda$) on memory samples while the true objective treats all samples with equal weight. This bias, introduced by ER's objective function, can contribute to the risk of memory overfitting.
\item \textbf{Problem-dependence}: given that $\beta_t\in (0,1]$, the biased weight on the memory samples is mostly influenced by $\lambda$. In other words, the memory overfitting risk is related to an inherent property of the CL problem concerned: the ratio $\lambda$ between the current task data size and the memory data size.\footnote{The experiments in this paper mainly consider a balanced CL setting where incoming tasks have the same data size. For imbalanced CL cases, $\lambda$ is also dependent on different tasks and should be expressed as $\lambda^\mathcal{T}$. The finding remains the same.}. 
While previous works extensively report empirical results on the influence of memory size, to our knowledge, we are the first to point out that the relative data size of an incoming task also plays an important role. Empirical evidence  is  provided in Section~\ref{sec:rar_hyper} to support this claim. With a 2k memory, performing rehearsal on the CORE50 dataset with a larger task data size ($\lambda=6$) faces a high level of memory overfitting while the CLRS dataset with a smaller task data size ($\lambda=1.12$) enjoys a lower risk of memory overfitting.
\item \textbf{Dynamic}: ER in online CL optimizes towards a \textit{dynamic} objective, which varies with each incoming batch $t$, as the weight on the memory sample depends on $N^t_{cur}$, and $N^{\mathcal{T}}_{past}$.
  This analysis shows that  memory overfitting may be relatively slight for the first few tasks when $N^{\mathcal{T}}_{past}$ is small. As more tasks arrive,  memory overfitting  worsens as $\beta_t$ increases with $N^{\mathcal{T}}_{past}$. In the case of an infinite data stream, $\lim_{N^{\mathcal{T}}_{past}\rightarrow \infty}\beta_t=1$, the memory weight is solely determined by $\lambda$.
\end{itemize}

\subsection{Repeated Rehearsal: The Decaying Regularization Effect of Incoming Data}
\label{sec:r-er}

We now investigate whether performing multiple iterations is beneficial to rehearsal or not. We refer to applying multiple iterations in ER as ``repeated experience replay'' (Repeated ER) or ``repeated rehearsal'' and formalize it as follows: for each incoming data batch $\mathcal{B}_t\sim \mathcal{D}_{\mathcal{T}}$ from the data stream, we perform multiple gradient updates ($K$ in total) using stochastic gradient descent (SGD) or variants thereof. At each gradient update $k = 1,...K$, a data batch $\mathcal{B}^\mathcal{M}_{t,k}$ is chosen from memory $\mathcal{M}$ and concatenated with the current incoming batch $B_t$ to perform a replay iteration as follows:
\begin{equation}
\label{eq:r-er}
  \theta_{t,{k+1}}=\theta_{t,k}-\frac{\eta}{\left|\mathcal{B}_{t}\right|} \sum_{\textbf{x},y \in  \mathcal{B}_t} \nabla \mathcal{L}\left(f_{\theta_{t,k}}(\textbf{x}),y\right) 
  -\frac{\eta}{\left|\mathcal{B}^\mathcal{M}_{t,k}\right|} \sum_{\textbf{x},y \in \mathcal{B}^\mathcal{M}_{t,k}} \nabla \mathcal{L}\left(f_{\theta_{t,k}}(\textbf{x}),y\right).
\end{equation}
Note that when $k=1$, this  update rule provides an unbiased stochastic gradient for Eq~\ref{eq:loss_er}; in the case of $k>1$, it provides a biased gradient estimate as the same incoming batch is used for consecutive gradient updates. To study the influence of this biased gradient update in repeated rehearsal, we examine the internal dynamics of repeated rehearsal by studying the loss landscape.

\begin{figure}[t]
\vskip -0.1in
\begin{center}
\subfigure[RER ($K=10$)]{
\includegraphics[width=0.49\columnwidth]{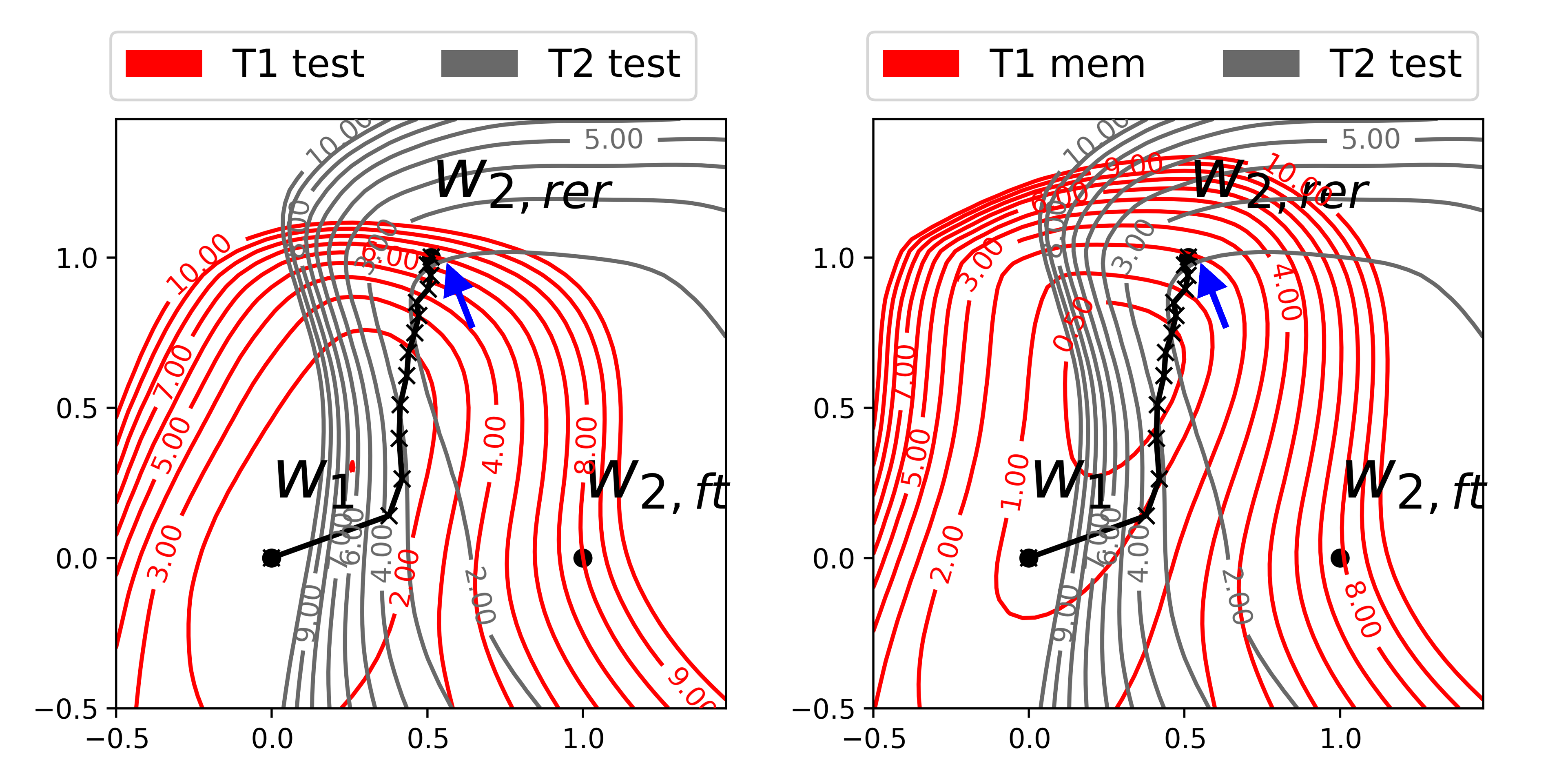}} 
\subfigure[RAR ($K=10$)]{
\includegraphics[width=0.49\columnwidth]{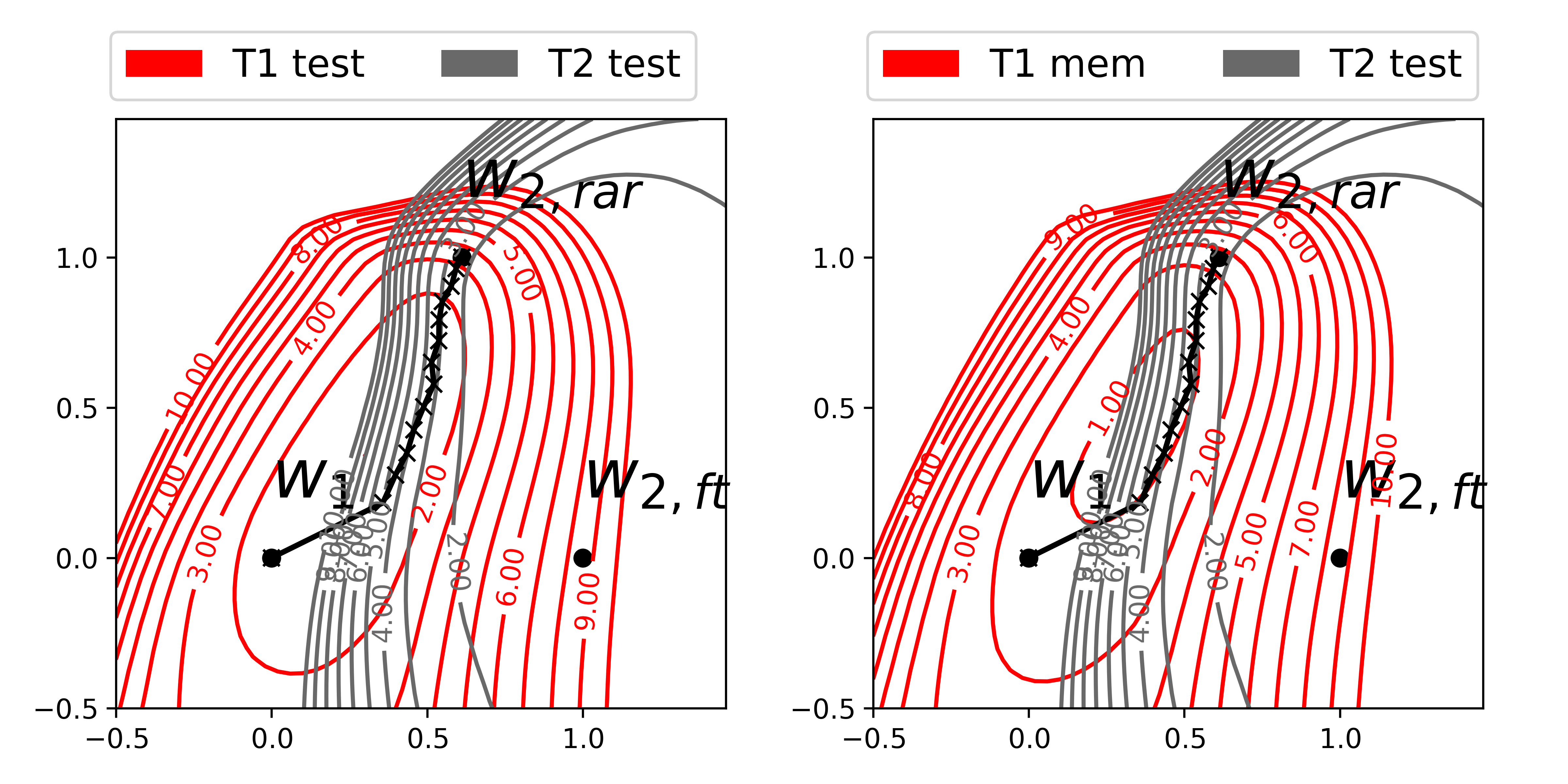}}
\caption{\color{black}Loss contours of RER and RAR. Memory data overfitting can be observed in (a) for RER but not in (b) for RAR.
Note how the shape and position of the loss contour of ``T1 test'' differs from the ``T1 memory'' loss contour in (a). At the CL solution point, the test loss is 7.9 (left blue arrow) while the memory loss is only 2.1 (right blue arrow).
}


\label{fig:loss_contour}
\end{center}
\vskip -0.2in
\end{figure}

\textbf{Memory Overfitting}: We compare the loss surfaces regarding the memory samples and the test data of past tasks during repeated rehearsal. Following the visualization method used in~\cite{verwimp2021rehearsal,mirzadeh2020linear}, we examine the learning process on the first two tasks in the Split Mini-ImageNet dataset and plot the loss landscape in the 2D plane defined by three model parameter vectors\footnote{When training $w_2$ and $w_{2,ft}$, the model is initialized from $w_1$. The memory contains 100 samples/task (see Fig~\ref{fig:loss_contour} (b) left and right). The 2-d coordinate system is built by orthogonalizing $w_2-w_1$ and $w_{2,ft}-w_1$.}: the model $w_1$ obtained by training on the first task until convergence,  the  model $w_2$ obtained on the second task using experience replay, and the  model $w_{2,ft}$ obtained after training on the second task using finetuning without replay. Verwimp et al. (\citeyear{verwimp2021rehearsal}) use this method to demonstrate the memory overfitting in  ER  for \textit{offline} continual learning setting with 10 epochs. Our results show that applying ER in the \textit{online} setting 
yields severe memory overfitting with 10 iterations (see Fig~\ref{fig:loss_contour} (a)). In other words, increasing the number of iterations means the loss landscape of the rehearsal memory provides a poorer approximation of the loss landscape of previous tasks, as shown by the differences in the positions and the shapes between the left and right red contours in Fig~\ref{fig:loss_contour} (a). 
As a result, the learning trajectory of RER avoids the high-loss ridge region for the memory data but goes right into the high-loss ridge region for the past tasks' test data.

\textbf{Regularization Effect of Incoming Data}: To investigate why repeated rehearsal may suffer from even more memory overfitting than vanilla rehearsal, we analyze the regularization effects of incoming data during repeated ER. To this end, we examine the training process given an incoming batch and compare the training loss on the memory batch and incoming batch with respect to memory iteration $k$ (see  Fig~\ref{fig:memloss_incomingloss} ). 
An interesting observation is that during the training session of a given incoming data batch, the decrease in the training loss on the incoming batch is much faster than the decrease in loss on the memory batches. One intuitive explanation is that the former is computed over a fixed batch during multiple iterations and
the latter is computed over different memory batch samples. As a result, even though the incoming loss is larger than the memory loss at the start  of a training session ($k=1$),
 at later iterations (i.e., $k>5$) the training loss of the incoming batch becomes $10-10^2$ times lower than that of the memory batch. This means that at this stage the regularization effect of the incoming data batch is greatly undermined: the joint training on the memory batch and the incoming batch becomes similar to training on memory only. 
 
In summary, our findings imply that the performance of online rehearsal is constrained by {\color{black} the dilemma between overfitting locally and underfitting globally}. Specifically, online rehearsal faces the challenge of underfitting of the large data stream but overfitting of a small memorized data subset. Applying repeated rehearsal ameliorates the former problem but aggravates the latter problem. Therefore, the performance gain from repeated rehearsal is quite limited.

\section{Repeated Augmented Rehearsal
}

\label{sec:rar}

\begin{figure}[t!]
\noindent\begin{minipage}{.5\textwidth}

\captionof{algorithm}{RL-based RAR}\label{algo_RAR}
\raggedright
\hrule 
 $\mathcal{M}$ is the memory with fixed size,\\
 $ \mathcal{B}_t$ is the incoming batch from the current task, \\
 $\theta$ are the parameters of the CL network, \\
 $w$ are the parameters of the RL agent, \\
 $K$ is the number of memory iterations, \\
 $P,Q$ are the augmentation hyperparameters\\
\begin{algorithmic}[1]
 \Procedure{RAR}{$\mathcal{M}_t$,  $ \mathcal{B}_t$, $\theta_t$, $w_t$ }
    \State $K_t,P_t,Q_t = SampleAction(w_t)$
    \For{$k = 1,...,K_t$}
      \State  $\mathcal{B}^\mathcal{M}_{t,k} \sim MemRetrieval(\mathcal{M}_t)$
      \State $\mathcal{B}_{aug} \leftarrow {\color{black}aug(\mathcal{B}^\mathcal{M}_{t,k} \cup \mathcal{B}_t,P_t,Q_t)}$
      \State$ r_t  \leftarrow ComputeReward(\mathcal{B}_{aug},\theta_{t,k})$
      \State $\theta_{t,k+1} \leftarrow SGD(\mathcal{B}_{aug},
\theta_{t,k})  $
    \EndFor\label{euclidendwhile}
    \State $\mathcal{M}_{t+1} \leftarrow MemUpdate(\mathcal{M}_t,\mathcal{B}_t) $
  \State $w_{t+1} \leftarrow UpdateRL(r_t)$
  \EndProcedure
  \hrule
\end{algorithmic}
\end{minipage}%
\begin{minipage}[t]{.5\textwidth}
\centering
\includegraphics[width=0.65\columnwidth]{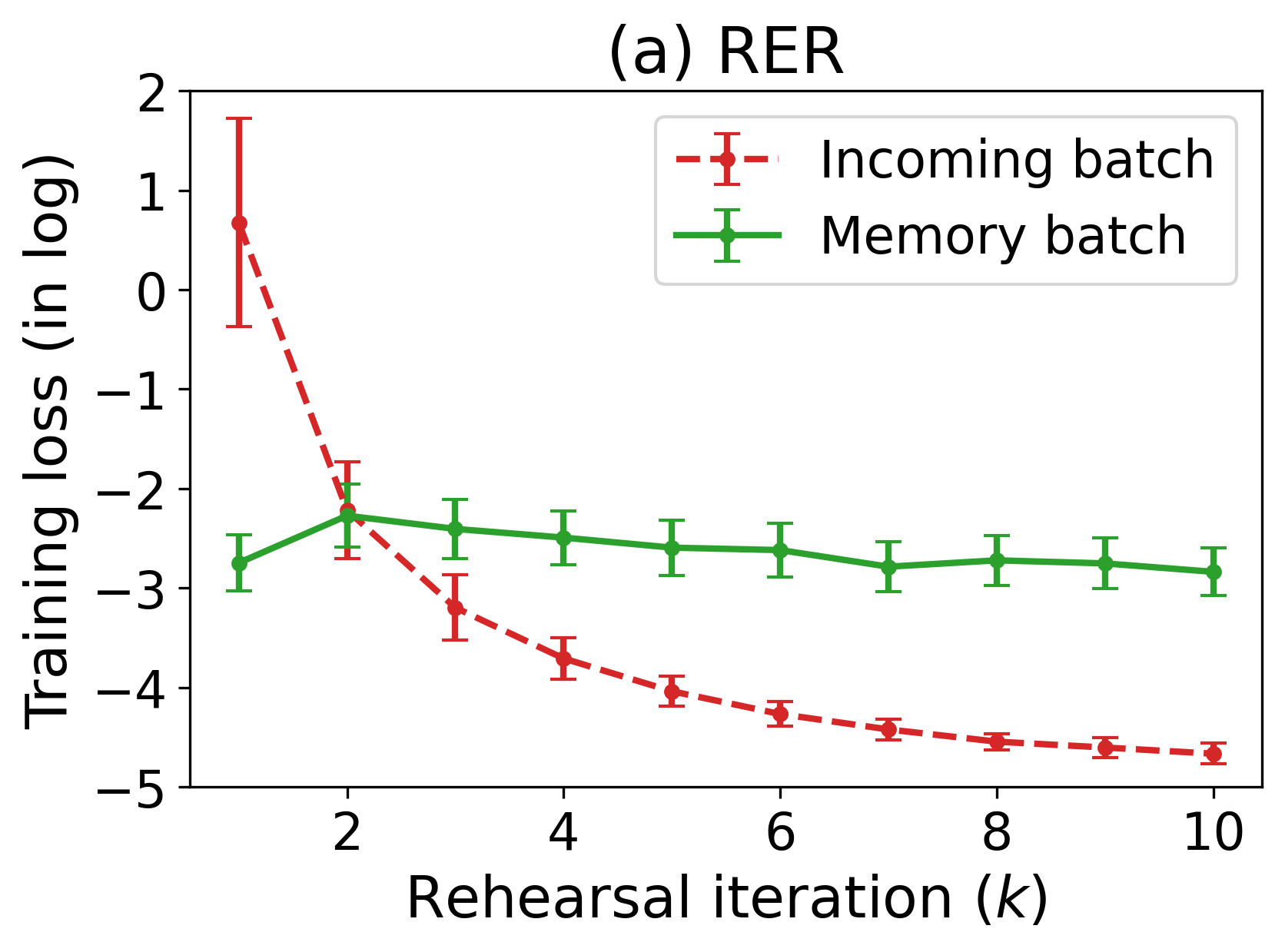}
\includegraphics[width=0.65\columnwidth]{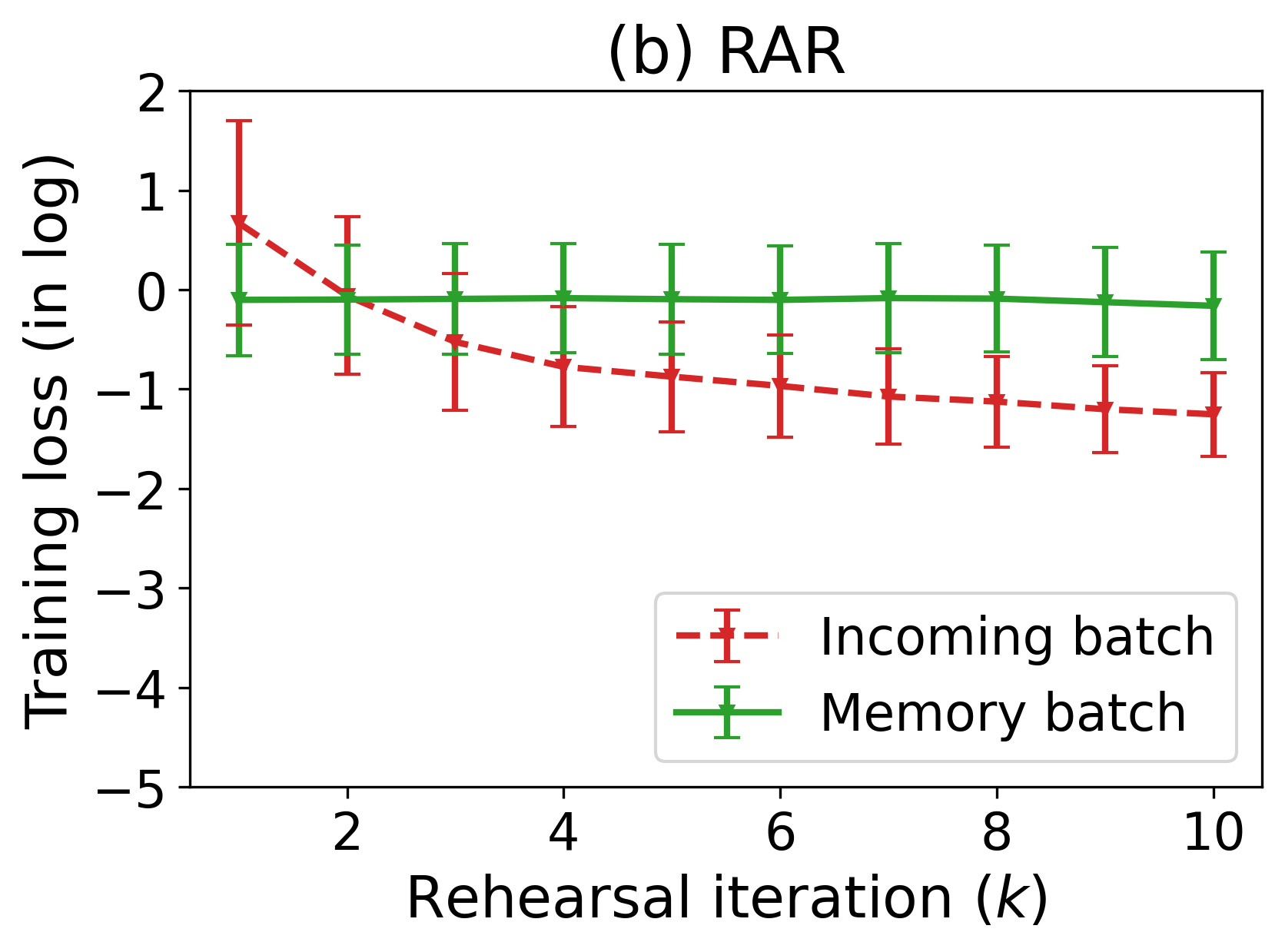}
\captionof{figure}{Memory loss vs. incoming loss.
}
\label{fig:memloss_incomingloss}
\vskip -0.3in
\end{minipage}
\end{figure}

 To deal with the overfitting-underfitting dilemma, we explore a simple strategy, ``repeated augmented rehearsal'' (RAR), which combines repeated rehearsal with data augmentation.   
Consider a group of transforms $G$ that acts on the input space $\mathcal{X}$  and is invariant under function $f$, i.e., $f(g\textbf{x})=f(\textbf{x}), g\in G, \textbf{x} \in \mathcal{X}$. Given an incoming batch from the data stream, $\mathcal{B}_t$, multiple replay iterations are conducted using this batch. At each replay iteration $k$, a random memory batch $\mathcal{B}^\mathcal{M}_{t,k}$ is sampled  and concatenated with the incoming batch. Then, a random transform $g_{t,k} \in G$ is sampled and applied to each data point $\textbf{x}_i$ in
the concatenated minibatch. The model parameters are updated as :
$$
g_{rar}=\frac{1}{\left|\mathcal{B}_{t}\right|} \sum_{\textbf{x},y \in  \mathcal{B}_t} \nabla \mathcal{L}\left(f_\theta(g_{t, k}\textbf{x}),y\right)
  +\frac{1}{\left|\mathcal{B}^\mathcal{M}_{t,k}\right|} \sum_{\textbf{x},y \in \mathcal{B}^\mathcal{M}_{t,k}} \nabla \mathcal{L}\left(f_\theta(g_{t, k}\textbf{x}),y\right)
.$$
Intuitively, the augmentation can help alleviate memory overfitting in two ways. First, we observe that applying augmentation on the incoming batch helps strengthen the regularization effect. As shown in Fig~\ref{fig:memloss_incomingloss}(b), the decaying regularization effect of {\color{black} incoming data} is alleviated in RAR, as the loss of the incoming batch stays comparable to the loss of the memory batch during multiple iterations.
Second, rehearsal on augmented memory batches can help to more accurately reflect past tasks' data distributions.
With RAR, the loss landscapes of memory data and past tasks' test data become very similar  (see Fig~\ref{fig:loss_contour} (b)), which suggests that the model ends up in a part of the parameter space where the rehearsal memory approximates the past tasks' distribution well. Moreover, the continual learning solution identified with RAR avoids the high-loss ridge not only in the memory data loss landscape but also in the test data loss landscape. 


Theoretically, we prove that augmented rehearsal reduces the generalization error in OCL.
 Specifically, assume the augmentation group $G$
 is a compact topological group and follows a probability distribution $\mathbb{Q}$. Similar to \textbf{Proposition 1}, it can be easily proven that the augmented rehearsal gradient corresponds to unbiased SGD on an augmented
empirical risk\footnote{Note that the theoretical analysis of the loss functions in Eq~\ref{eq:loss_er} and Eq~\ref{eq:loss_aer} is also applicable to offline continual learning, which may be of independent interest. More discussion of the influence of augmentation in offline continual learning can be found in Appendix~\ref{sec:offline_cl}.}  
(see \textbf{Proposition 3} in Appendix~\ref{sec:theory}):
\begin{equation}
\label{eq:loss_aer}
     \bar{\mathcal{R}}_t(\theta)=\sum_{\textbf{x},y\in \mathcal{D}_\mathcal{T}}\int_G\mathcal{L}(f_\theta(g\textbf{x}),y)d\mathbb{Q}(g) 
+\beta_t\lambda\sum_{\textbf{x},y \in \mathcal{D}_\mathcal{M}}\int_G\mathcal{L}(f_\theta(g\textbf{x}),y)d\mathbb{Q}(g).
\end{equation}
This result shows that applying augmented rehearsal is equivalent to performing an averaging operation of the loss of rehearsal in Eq~\ref{eq:loss_er} over the orbits of a certain group that keeps the data distribution approximately invariant. In the standard i.i.d. learning setting,  \cite{chen2020group} found that such an orbit-averaging operation can reduce both the variance and generalization error. Based on Eq~\ref{eq:loss_er} and Eq~\ref{eq:loss_aer}, we show that this theoretical benefit of using augmentation to boost model invariance is also applicable to rehearsal in continual learning. In fact, as discussed in Section~\ref{sec:erm}, rehearsal in online CL has a biased empirical risk Eq~\ref{eq:loss_er}, which leads to inherent memory overfitting and poor generalization ability. Thus, this benefit of augmentation in reducing generalization error is particularly important when applying rehearsal in online CL. 


 The modifications required for the RAR procedure are summarized in Lines 3 and 5 of Algorithm~\ref{algo_RAR}. It uses a general framework for ER-based continual learning that consists of three key components: sampling from memory, joint training on memory data and incoming data, and updating of the memory. As mentioned in the related work section, different ER variants have been proposed to improve these components. 
RAR can flexibly be combined with any of these ER variants, and we investigate the effectiveness of RAR on these different ER variants in the experiment section.


\section{Reinforcement Learning-based Adaptive Repeated Augmented Rehearsal}
There are two key components in RAR: repeated rehearsal and augmented rehearsal. The interplay of the two is determined by the number of memory iterations and the strength of augmentation. A key question is how to choose these hyperparameters. In general, hyperparameter tuning (HPT) still remains an unsolved challenge for online CL due to the single-pass assumption~\citep{chaudhry2018efficient}. Finding suitable RAR hyperparameters needs to account for the severity of memory overfitting and poses extra challenges. In particular, as shown in the ERM analysis, the extent of memory overfitting is related to the CL problem features (e.g., task data size and memory size) and also varies at different training stages of the continual learning process. To automatically select suitable RAR hyperparameters for the different CL problems and different training stages, we propose to use reinforcement learning to adaptively adjust the hyperparameters (see Algorithm~\ref{algo_RAR}). 

In particular, we design the hyperparameters of RAR as the action space and use the training statistics as the reward (see lines 2, 6, and 10 in Algorithm~\ref{algo_RAR}). A major challenge of applying RL in online HPT is sample efficiency. The exploration horizon (i.e, training steps) in the OCL environment is quite limited due to the constraint of a single pass through the data and  poor action choices (undesirable hyperparameters) may lead to a bad gradient update step and hurt the OCL training process. To address the sample efficiency issue, we employ the multi-armed bandit framework and apply bootstrapped policy gradient (BPG)~\citep{zhang2019bootstrapped}. The key idea of BPG is to incorporate prior knowledge to bootstrap the policy gradient to achieve stable and fast convergence with limited samples. To obtain prior knowledge in OCL problems, 
we use the training accuracy on the memory batch as the overfitting feedback, as a higher training memory accuracy suggests a higher chance of memory overfitting.
Reward is defined as the distance between the current memory accuracy and target memory accuracy. Compared against a target memory accuracy (e.g., 0.9), the current memory accuracy is used to indicate whether the current choice of rehearsal iteration or augmentation causes too much memory overfitting
and then the action selection probability is adjusted following BPG (see Appendix~\ref{sec:rl_implementation} for more RL design and implementation details). The algorithm details of applying BPG as a specific RL method for hyperparameter tuning is summarized in  Algorithm~\ref{algo_RAR_BPG} of  Appendix~\ref{sec:rl_implementation}).

\begin{table*}[t]
\caption{Accuracy on four OCL benchmarks with 2k and 5k memory. The performance boost of RAR over ER and ER variants is shown.
} 
\label{tab:main}
\begin{center}
\begin{small}
\begin{sc}

\resizebox{1.0\textwidth}{!}{%
  \begin{tabular}{l|cc|cc|cc|cc}
\toprule
   &
      \multicolumn{2}{|c|}{Seq-CIFAR100} &
      \multicolumn{2}{|c|}{Seq-MINI-IMAGENET} &
      \multicolumn{2}{|c}{CORE50-NC}  &
           \multicolumn{2}{|c}{CLRS25-NC}  \\
    & 2K & 5K & 2K & 5K & 2K & 5K & 2K & 5K\\
\midrule
Finetune & \multicolumn{2}{c|}{3.2 $\pm$ 0.1} & \multicolumn{2}{c|}{4.3 $\pm$ 0.8}  & \multicolumn{2}{c}{7.7 $\pm$ 0.2} &
\multicolumn{2}{|c}{6.5 $\pm$ 0.9} \\
LWF & \multicolumn{2}{c|}{8.7 $\pm$ 0.5} &\multicolumn{2}{c|}{10.9 $\pm$ 0.5 }&  \multicolumn{2}{c}{9.6 $\pm$ 0.3} & \multicolumn{2}{|c}{ 12.4 $\pm$ 2.2} \\
AGEM & 8.5 $\pm$ 0.4 & 9.2 $\pm$ 0.2&11.6 $\pm$  0.1&	13.1 $\pm$  0.3 &	18.6 $\pm$ 0.4&	19.4 $\pm$ 1.8 &	14.6 $\pm$  1.4 &	14.4 $\pm$ 0.3
\\ 
 \hline
\midrule
ER & 19.0 $\pm$ 0.6 & 26.2 $\pm$ 0.2 &20.0 $\pm$ 0.8 & 23.0 $\pm$ 0.6& 
24.0 $\pm$ 2.0
&27.8 $\pm$ 0.2


& 18.7 $\pm$ 1.6 & 19.2 $\pm$0.3 
 \\
ER-RAR & 27.8 $\pm$ 0.5 & 36.2 $\pm$ 0.7 &30.0 $\pm$ 0.9  &36.5 $\pm$ 0.4  & 39.3 $\pm$ 1.4 & 45.0 $\pm$ 2.7 & 28.6 $\pm$ 2.7 &28.9 $\pm$	1.5
\\
Gains &8.8 $\uparrow$
&10.0 $\uparrow$&10.0$ \uparrow$&
13.5 $\uparrow$ & 15.3 $\uparrow$ & 17.2 $\uparrow$ & 9.9 $\uparrow$ & 9.7 $\uparrow$ \\
\midrule
MIR &18.4 $\pm$ 0.8 & 25.7 $\pm$ 1.8 & 19.4 $\pm$ 0.6 &22.3 $\pm$ 0.2  & 25.2 $\pm$ 1.3 & 26.9 $\pm$ 0.9& 14.3 $\pm$ 3.6 &15.2 $\pm$	3.0
\\
MIR-RAR & 27.5 $\pm$ 0.2& 36.1 $\pm$ 0.3 & 29.5 $\pm$ 0.6 & 34.9 $\pm$ 0.7  & 39.1 $\pm$ 1.0 &44.6 $\pm$ 1.7& 27.8 $\pm$ 1.6 &29.2 $\pm$	2.6
\\
Gains&9.1 $\uparrow$ &10.4 $\uparrow$ & 10.1 $\uparrow$ &12.6 $\uparrow$ & 13.9 $\uparrow$ & 17.7 $\uparrow$ & 13.5 $\uparrow$ & 14.0 $\uparrow$\\
\midrule
ASER &20.9 $\pm$ 0.3 & 24.3 $\pm$ 2.0 & 15.7 $\pm$ 0.1 & 17.5 $\pm$ 0.7  & 16.4 $\pm$ 1.4 & 16.7 $\pm$ 2.3 & 19.4 $\pm$ 1.3  &19.7 $\pm$	1.4
\\
ASER-RAR &28.1 $\pm$ 0.3 & 35.8 $\pm$ 1.0 & 27.0 $\pm$ 0.3 & 32.2 $\pm$ 0.6  & 24.2 $\pm$ 0.4 & 30.0 $\pm$ 1.6 & 28.7 $\pm$ 0.2 &29.5 $\pm$	0.2 
\\
Gains&7.2 $\uparrow$&11.5 $\uparrow$ &11.3 $\uparrow$ & 14.7 $\uparrow$  & 7.8 $\uparrow$ & 13.3 $\uparrow$ & 9.3 $\uparrow$ & 9.8 $\uparrow$\\
\midrule
SCR &32.0 $\pm$ 1.1 & 37.4 $\pm$ 0.2 &29.7 $\pm$ 1.0 &33.1 $\pm$ 1.9  & 45.1 $\pm$ 0.1 & 50.3 $\pm$ 1.9 & 23.5 $\pm$ 2.2 & 23.6 $\pm$	3.0
\\
SCR-RAR & {37.1 $\pm$ 0.7 }& {45.8 $\pm$ 0.2 }& {35.4 $\pm$ 0.7} &{43.7 $\pm$ 0.4 } &{53.4 $\pm$ 0.9} & {61.1 $\pm$ 1.1} & {37.4 $\pm$ 1.0} & {41.5 $\pm$	0.9} \\
Gains & 5.1 $\uparrow$ &8.4 $\uparrow$ &5.7 $\uparrow$ &10.6 $\uparrow$ &8.3 $\uparrow$ &10.8 $\uparrow$ &14.9 $\uparrow$ &17.9 $\uparrow$ \\
 \midrule
{\color{black}ER\tiny{rw}} &  21.0 $\pm$ 1.0 & 26.2 $\pm$	0.2
&20.1 $\pm$ 0.8  & 23.0	$\pm$ 0.6 
& 24.6 $\pm$ 0.6 & 27.8 $\pm$	0.8
& 19.2 $\pm$0.6 & 19.2 $\pm$	0.3 \\
{\color{black}ER{\tiny{rw}}-RAR}
& 30.8 $\pm$ 0.1 &36.5  $\pm$	0.4
& 30.4 $\pm$ 1.3 &36.5 $\pm$	0.4
& 45.3 $\pm$ 2.2 &50.8  $\pm$	0.9
& 28.6 $\pm$ 2.7 &28.9  $\pm$	1.5 \\
Gains & 9.8 $\uparrow$ & 10.3 $\uparrow$ & 10.3 $\uparrow$ &13.5 $\uparrow$ &20.7 $\uparrow$ &23.0 $\uparrow$  & 9.4 $\uparrow$ &9.7 $\uparrow$
 \\ \midrule
{\color{black} DER} & 8.4  {$\pm$ 0.6} & 9.1 $\pm$ 0.3 & 11.8 {$\pm$ 0.5} & 12.3 $\pm$ 1.7 & 23.8 $\pm$ 0.6 & 23.4 $\pm$ 2.5 & 11.8 {$\pm$ 2.6} & 12.6 $\pm$ 1.1\\
{\color{black}DER-RAR} & 30.0 {$\pm$ 1.2} & 41.9 $\pm$ 0.5 & 26.2 {$\pm$ 0.4} &35.5 $\pm$ 1.5 &37.7 $\pm$ 1.4 & 42.0 $\pm$ 3.7 & 28.4 {$\pm$ 3.2} & 27.4 $\pm$ 3.8 \\
 Gains & 21.6 $\uparrow$ & 32.8 $\uparrow$ & 14.4  $\uparrow$ &23.2 $\uparrow$  &11.9  $\uparrow$ & 18.6 $\uparrow$ & 16.6 $\uparrow$ &  14.8$\uparrow$\\
\bottomrule
\end{tabular}}
\end{sc}
\end{small}
\end{center}
\end{table*}

\section{Experiments}

\subsection{Experiment Setup}
\label{sec:setup}
\textbf{Baseline}: We apply RAR to four ER-based continual learning algorithms:
ER~\citep{chaudhry2019tiny}, MIR~\citep{aljundi2019online}, ASER~\citep{shim2021online}, and SCR~\citep{mai2021supervised}. 
We also compare it with other continual learning methods, including the regularization-based  method LWF~\citep{li2017learning} and the constrained optimization-based method A-GEM~\citep{chaudhry2018efficient}.

\textbf{Dataset}: Four CL benchmarks are used in the experiments: 
Seq-CIFAR100 (20 tasks),  Seq-MiniImageNet (10 tasks)~\cite{vinyals2016matching}, CORE50-NC (9 tasks) ~\cite{lomonaco2017core50} and CLRS25-NC (5 tasks)~\cite{li2020clrs}(see Appendix~\ref{sec:dataset} for more details). Additionally, we also investigate ER and RAR on the large-scale ImageNet-1k dataset in Appendix~\ref{sec:imagenet}.

\textbf{Implementation}: We use a reduced ResNet-18 for all datasets following~\cite{mai2021supervised,aljundi2019online}. Single-head evaluation is employed with a shared final layer trained for all the tasks.
RandAugmentation~\cite{cubuk2020randaugment} is used for auto augmentation. Given a set of augmentation operations, it randomly selects $P$ augmentation operations and exerts an augmentation magnitude of $Q$ for all the selected augmentation operations on each image. All the experimental results we present are averages  of three runs. We summarize all hyperparameter details in Appendix~\ref{sec:hyper}.
The running time of different algorithms is shown in Appendix~\ref{sec:running_time}. 


\subsection{Main Results}
\label{sec:main_res}
\begin{figure}[]
\begin{center}
\includegraphics[width=0.45\columnwidth]{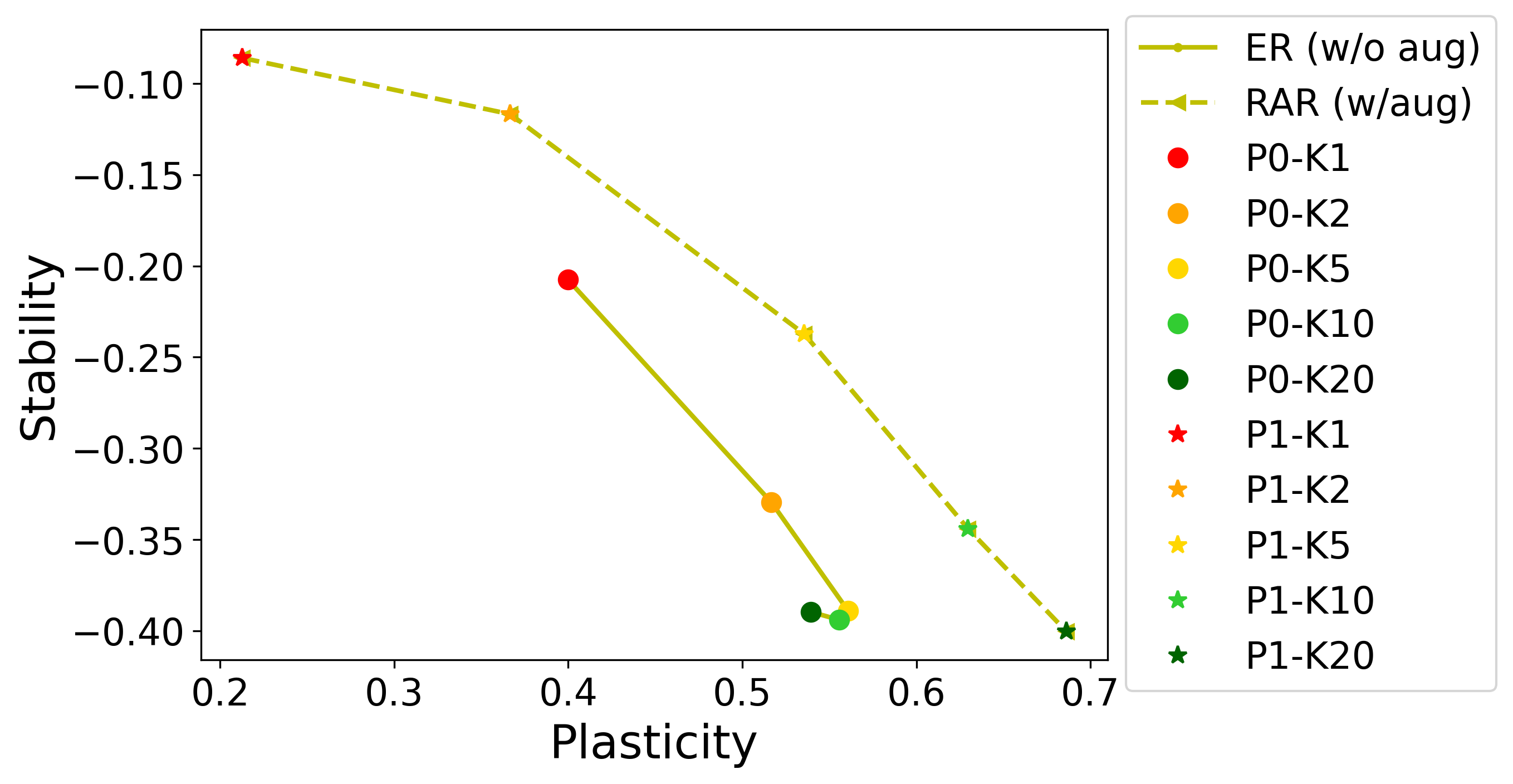}
\includegraphics[width=0.45\columnwidth]{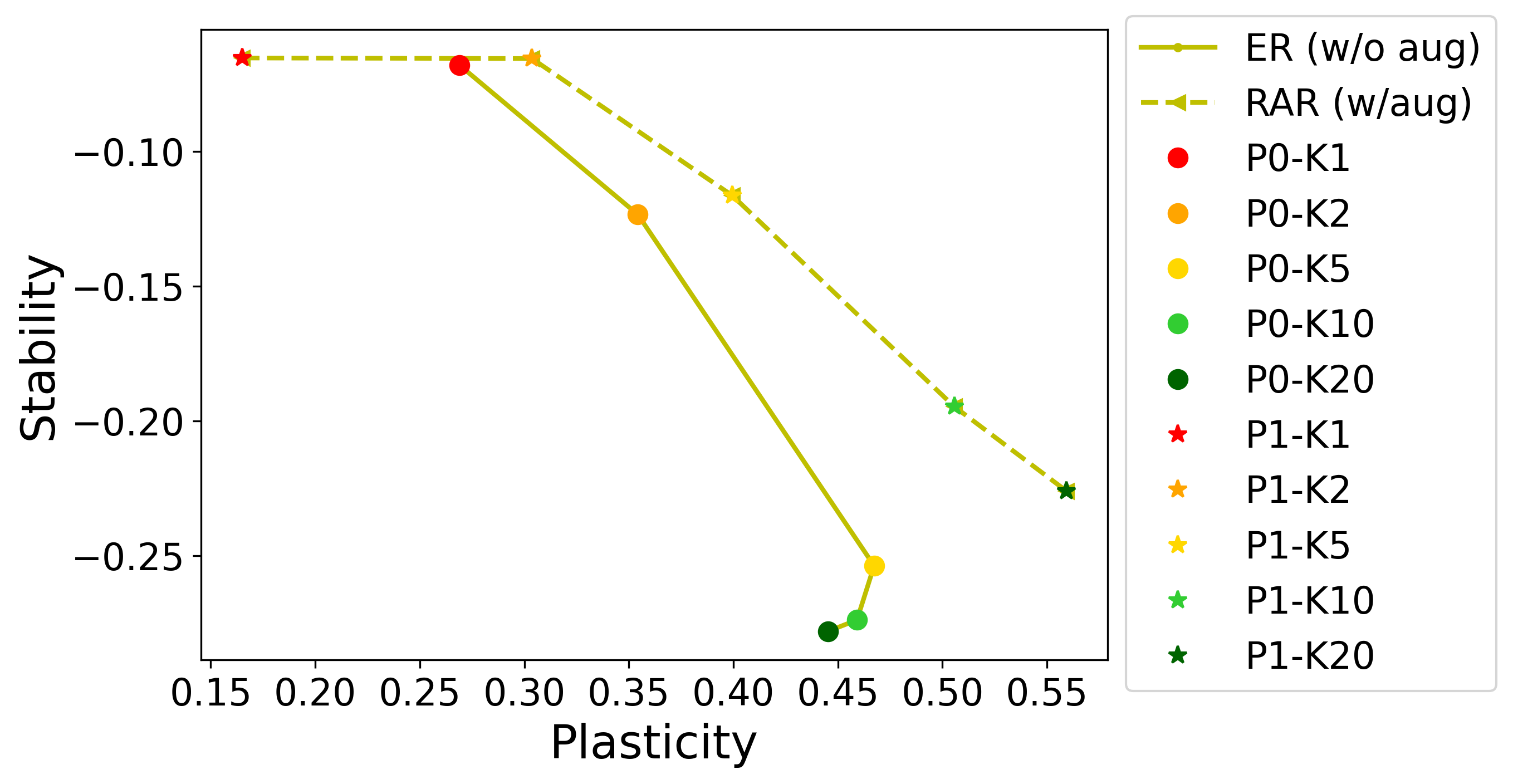}
\caption{Stability and plasticity trade-off: CIFAR100 (left) and Mini-ImageNet(right)}
\label{fig:forgetting}
\end{center}
\vskip -0.3in
\end{figure}
\textbf{RAR with ER and its variants} We first analyze RAR's performance with a pre-defined hyperparameter set ($K=10,P=1,Q=14$). As shown in Table~\ref{tab:main}, RAR greatly improves the ER method on the four datasets, by $+8.8\% \sim +17.2\%$. Moreover, RAR also leads to substantial gains for the other algorithmic variants of ER for all datasets (MIR: $+9.1\% \sim +17.7\%$, ASER: $+7.2\% \sim +14.7\%$, SCR: $+5.1\% \sim +17.9\%$). These results suggest that even with advanced memory management strategies, such as MIR or ASER, or representation learning techniques, e.g., SCR, OCL still benefits substantially from repeated augmented rehearsal. 

{\color{black}
\textbf{RAR with Modified Rehearsal Loss } Besides using the vanilla online rehearsal loss in Eq~\ref{eq:loss_er}, we investigate the effectiveness of RAR with another two, more advanced, rehearsal loss designs: 1) \textbf{Reweighted memory loss}: ER-rw introduces a reweighting hyperparameter $\alpha$ in the gradient of Eq~\ref{eq:grad_er} to deal with the biased ER loss by balancing the weight of the memory loss and incoming loss;
 2) \textbf{Distillation-based memory loss}: DER~\citep{buzzega2020dark} employs the logits-based distillation loss for memory samples, instead of the cross entropy loss. The results in Table 1 show RAR leads to large performance gains for ER-rw (for the best $\alpha$ choice; further results can be found in Appendix~\ref{sec:reweighted_ER}) and DER for all four datasets (ER-rw:$+9.4\% \sim +23.0\%$, DER:$+11.9\%\sim+32.8\%$).
 This suggests that even with more advanced rehearsal loss designs, repeated augmented rehearsal is important for online rehearsal. 

}

\textbf{Stability and Plasticity Trade-off} Based on the definition of accuracy $A_T$, forgetting $F_T$ and backward transfer $B_T$ in Section~\ref{sec:related_work_OCL}, we find that these three metrics have the following relationship:
\[{A_T}=\underbrace{\frac{1}{T} \Sigma_{i=1}^{T} a_{i, i}}_\text{Plasticity}+\underbrace{\frac{T-1}{T} {B_T}}_\text{Stability} \ge \frac{1}{T} \Sigma_{i=1}^{T} a_{i, i}-\frac{T-1}{T} {F_T}.\]
Interestingly, this finding shows that accuracy is related to the ability to learn new tasks, quantified by $\frac{1}{T} \sum_{i=1}^{T} a_{i, i}$, and the ability to avoid forgetting past tasks, quantified by $\frac{T-1}{T}{B_T}$, which draws a connection to the more general problem of the stability-plasticity trade-off in neural networks and continual learning~\citep{grossberg2012studies,delange2021continual}. Plasticity refers to the ability to integrate new knowledge and  stability refers to the ability to retain old knowledge. Fig~\ref{fig:forgetting} presents the stability and plasticity trade-off in RAR. Generally, we observe increasing the repeated rehearsal iterations ($K$) leads to a higher level of  plasticity. However, this may also cause a decrease of stability, i.e., introduce forgetting. On the other hand, the use of augmentation generally improves stability. More importantly, the use of augmentation in repeated rehearsal shifts the stability-plasticity trade-off curve towards the upper right, thus creating a better stability-plasticity trade-off frontier. 

 \textbf{Hyperparameter Tuning for RAR} We compare the RL-based hyperparameter tuning method with the hyperparameter tuning framework for continual learning (HTOCL) method used in \cite{chaudhry2018efficient,mai2022online} (see Section~\ref{sec:related_work_OCL}). 
The results in Table~\ref{tab:rar_variants} show that RL-based RAR significantly outperforms using HTOCL to select hyperparameters for RAR. One reason is that the extent of memory overfitting varies at the different training stages of CL. HTOCL only uses the first few tasks to select hyperparameters for RAR, which may not optimal for later stages of CL. In fact, we observe HTOCL tends to select a large number of repeats and small augmentation strength. This selection strategy may be desirable for  problems with short task sequences but problematic for long task sequences  with increased memory overfitting risk. In contrast, the RL-based method can take into account the latest feedback (e.g., train memory accuracy) to adjust hyperparameter choices. The selected iteration numbers and augmentation are shown in Appendix~\ref{sec:action_selection}.

\begin{table}[]
\caption{Accuracy of variants of RAR and different hyperparameter tuning methods. } 
\label{tab:mem_aug}
\begin{center}
\begin{small}
\begin{sc}
\resizebox{0.8\textwidth}{!}{%
\begin{tabular}{l|cccc}
\toprule
 & Seq-CIFAR100 & Seq-Mini-ImageNet  & CORE50-NC & CLRS25-NC\\
\midrule
ER      &19.0 \tiny{$\pm$ 0.6}		&20.0 \tiny{$\pm$	0.8}		&24.0 \tiny{$\pm$	2.0} &18.7 \tiny{$\pm$ 	1.6} \\ 
RAR\tiny{-mem} &25.4 \tiny{$\pm$	0.7}		&27.4 \tiny{$\pm$	0.8}		&38.6 \tiny{$\pm$	0.7} & 28.8 \tiny{$\pm$	1.0}
 \\
RAR\tiny{-inc} &21.6 \tiny{$\pm$	0.2}		&24.5 \tiny{$\pm$	0.1}		&35.7 \tiny{$\pm$	1.1} & {29.1 \tiny{$\pm$	1.1}} \\
RAR\tiny{-both}     &{27.8 \tiny{$\pm$ 0.5}}		&{30.0 \tiny{$\pm$	0.9}}		& {39.3 \tiny{$\pm$	1.4}} &28.6 \tiny{$\pm$		2.7} \\ \midrule 
RAR-HTOCL 
& 23.4 \tiny{$\pm$ 0.2}
& 26.0 \tiny{$\pm$ 0.2}
& 40.8 \tiny{$\pm$ 0.7}
& 26.9 \tiny{$\pm$ 0.5}
 \\
RAR-RL
& \textbf{29.6 \tiny{$\pm$ 0.4}}
& \textbf{32.1 \tiny{$\pm$ 1.0}}
& \textbf{44.4 \tiny{$\pm$ 0.8}}
& \textbf{35.0 \tiny{$\pm$ 0.7}} \\
\bottomrule
\end{tabular}
}
\end{sc}
\end{small}
\end{center}
\label{tab:rar_variants}
\end{table}

\subsection{Ablation Studies}
\label{sec:ablation}

\label{sec:rar_hyper}
\begin{figure*}[t]
\begin{center}
\subfigure[CIFAR100]{
\includegraphics[width=0.23\columnwidth]{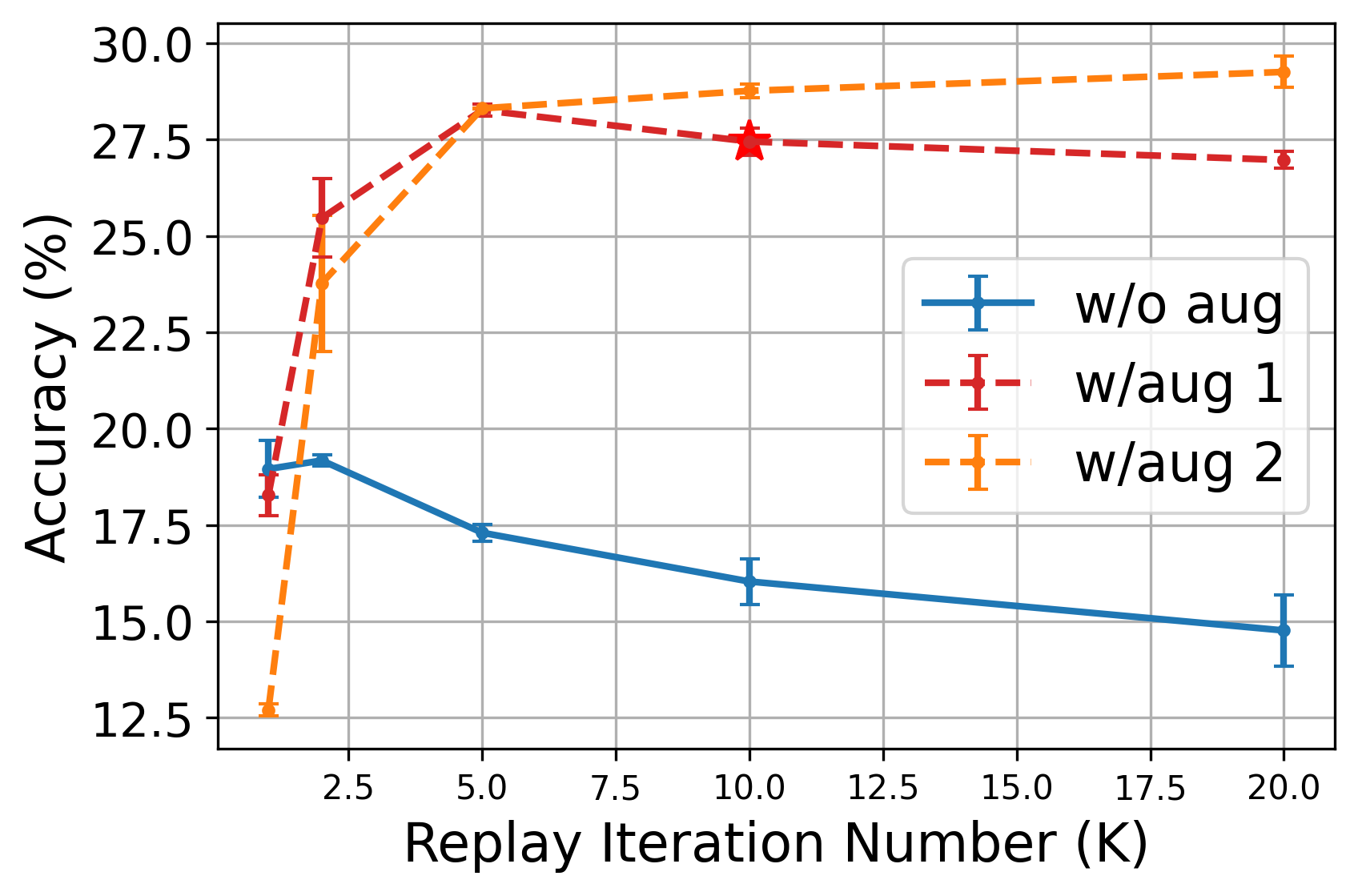}}
\subfigure[MINI-IMAGENET]{
\includegraphics[width=0.23\columnwidth]{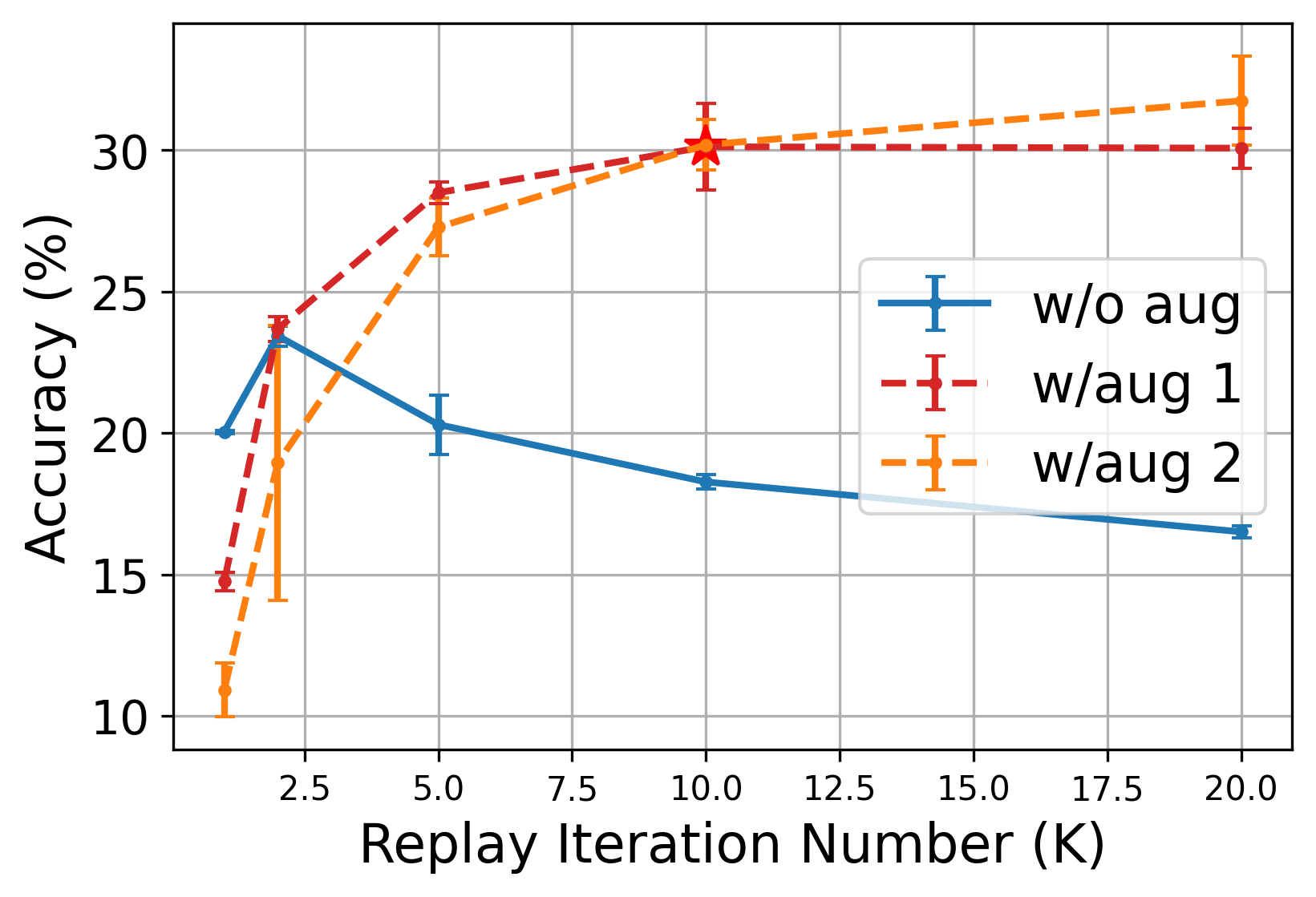}}
\subfigure[CORE50]{
\includegraphics[width=0.23\columnwidth]{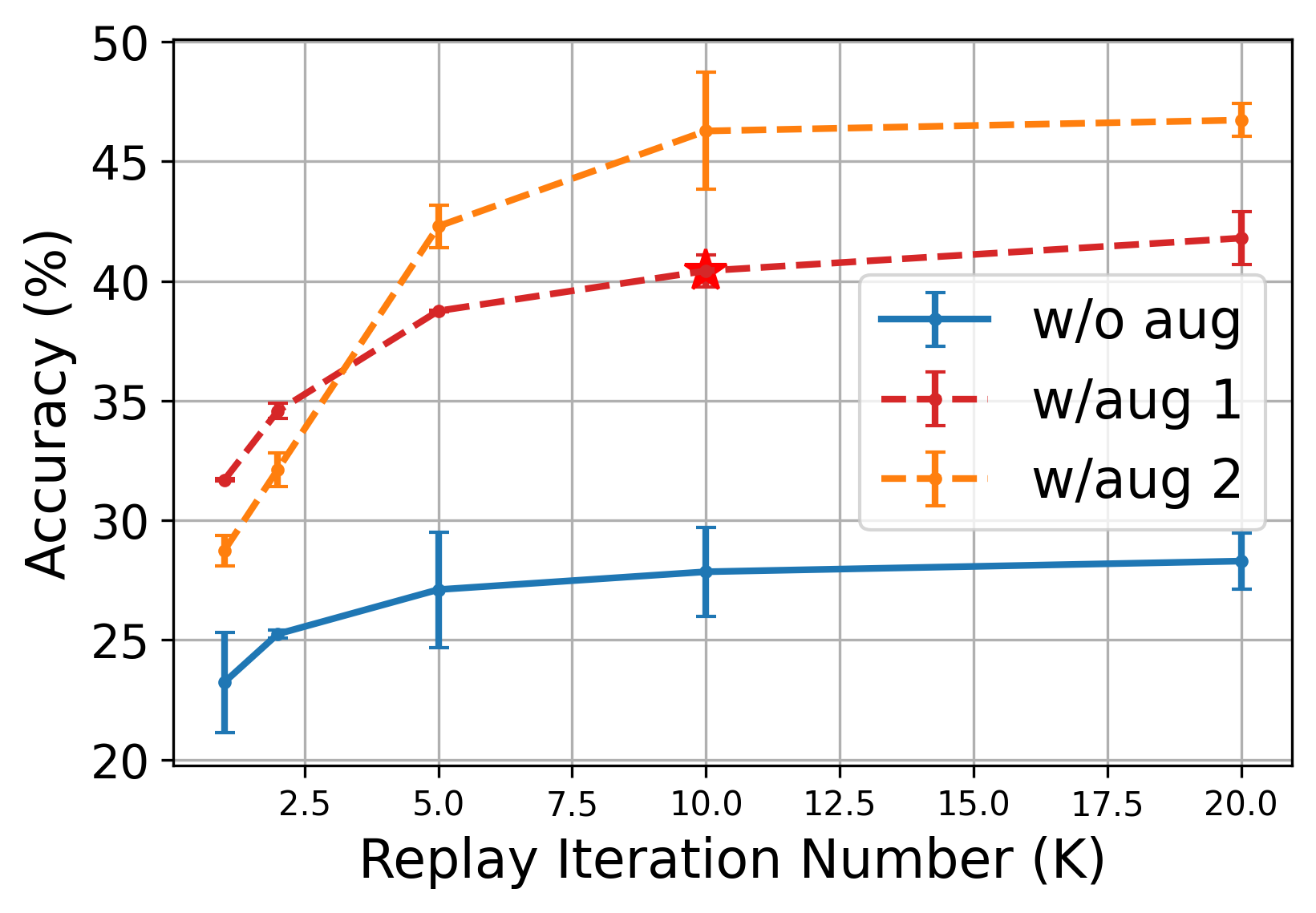}}
\subfigure[CLRS25]{
\includegraphics[width=0.23\columnwidth]{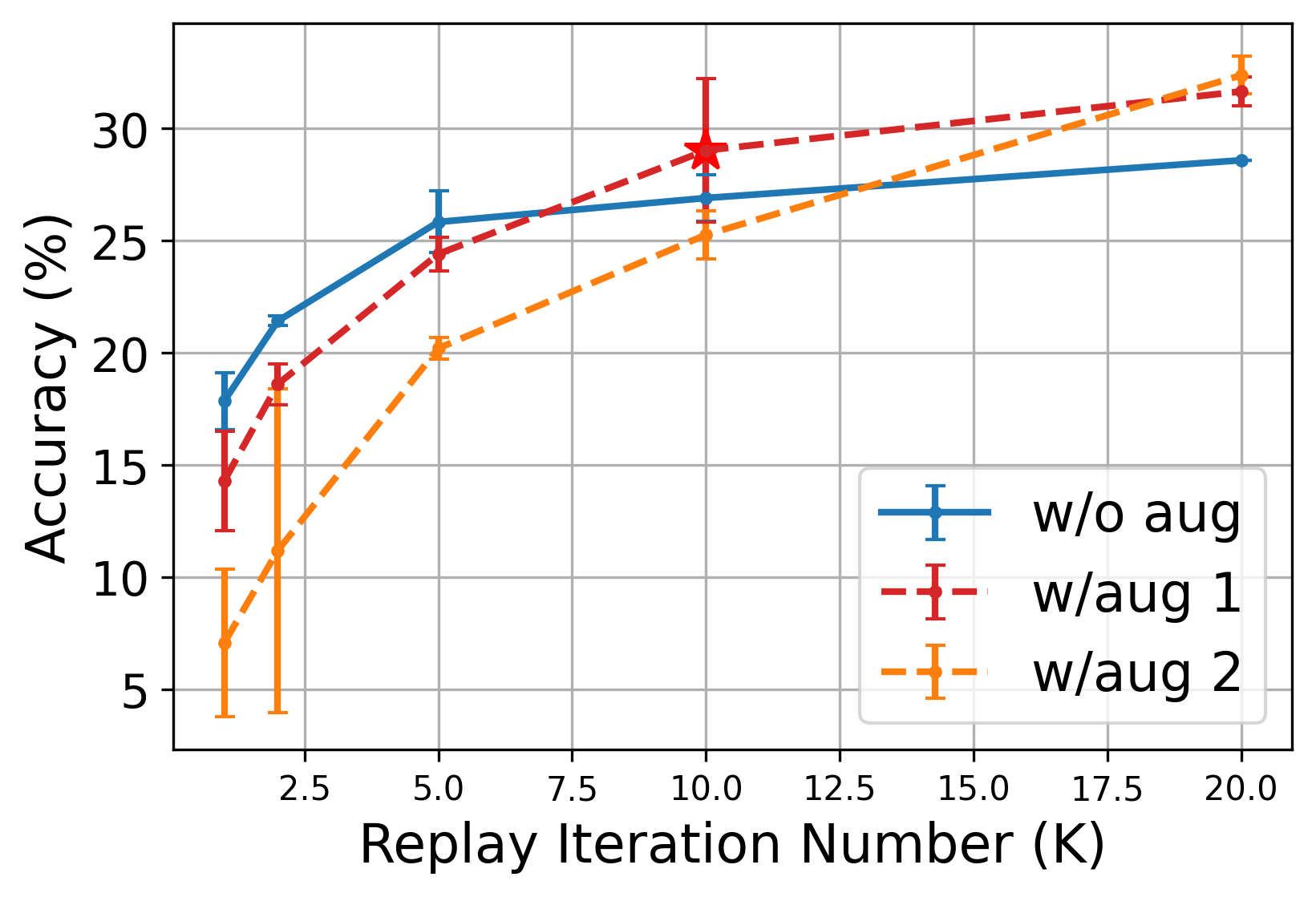}}
\caption{Effects of augmentation and rehearsal iterations (red stars: accuracy in Table~\ref{tab:main}).
}
\label{fig:aug_para}
\end{center}
\vskip -0.2in
\end{figure*}

\textbf{Interplay between Repeated and Augmented Rehearsal} We investigate the interaction of the number of replay iterations with the augmentation strength in Fig~\ref{fig:aug_para}. 
For three out of four datasets, using augmentation alone without repeated rehearsal leads to even worse performances than rehearsal without augmentation (see $K=1$ in Fig~\ref{fig:aug_para}). One explanation is the underfitting challenge of online CL. Training on augmented samples can make model underfitting even worse. The only exception is the CORE50 dataset, which inherently has a high memory overfitting risk with $\lambda=6$ making it benefit more from augmentation. Similarly, employing repeated rehearsal alone (see the blue solid line in Fig~\ref{fig:aug_para}) also harms performance in three out of four datasets, with the CLRS dataset as the only exception, which enjoys a low memory overfitting risk with $\lambda=1.12$. An important takeaway message is that repeated rehearsal or augmented rehearsal is not always helpful in OCL settings and whether they will benefit or harm the performance is dependent on the structure of the OCL problem at  hand (e.g., the task data size and memory data size). 

 \textbf{RAR's Robustness to Large Numbers of Repeats} Although the performance curve flattens out around 10 iterations, there is no evident drop in performance even with 20 iterations. This result reinforces how RAR can help with the underfitting-overfitting dilemma: it can support the use of more training in OCL without having to worry about the performance drop introduced by memory overfitting.
In comparison, for repeated rehearsal without augmentation (solid lines), the accuracy starts to drop quickly on CIFAR100 and MiniImageNet when using more than two iterations.

\textbf{Augmenting the Memory vs. Augmenting the Incoming Data} RAR applies augmentation to both the memory batch and the incoming batch. We examine the effectiveness of the two separately. Table~\ref{tab:mem_aug} presents the performance of RAR applied (a) solely with memory augmentation (RAR-mem) and (b) solely with incoming data augmentation (RAR-inc). 
We find that RAR achieves the best performance compared to RAR-mem and RAR-inc. This shows that it is beneficial to apply augmentation to both the memory batch and the incoming  batch. Interestingly,  RAR-inc itself also achieves consistent performance improvements over RER, e.g., a gain of 7.7\% in CORE50. 
As discussed in Section~\ref{sec:algo}, adding augmentation on the incoming batch can strengthen the regularization effect of the incoming task and indirectly alleviate memory overfitting. 




\textbf{RAR with MIR, ASER, SCR} We also perform ablation studies to investigate RAR's strong performance with the ER variants (see Appendix~\ref{sec:RAR_SCR}). Similar to ER, the performance gains for RAR-MIR and RAR-ASER come from the combination of repeated rehearsal and augmented rehearsal. However, for SCR, the repeated rehearsal itself also leads to a consistent performance boost. One reason is that SCR already includes a strong augmentation procedure with four augmentation operations to construct the supervised contrastive loss used in this method. It also works poorly without augmentation.

{\color{black}\textbf{Augmentation for Offline Rehearsal} Although this paper focuses on online CL, the (augmented) empirical risk analysis is also relevant to offline CL (see Propositions 2 and 3 in Appendix~\ref{sec:theory}). This suggests the memory overfitting risk in offline rehearsal is related to the ratio of task-to-memory size $\lambda$ and can be alleviated by augmented rehearsal (see more empirical results in Table~\ref{tab:offline_cl} of Appendix~\ref{sec:offline_cl}).
}

\section{Discussion and Conclusion}

Rehearsal-based methods play a central role in fighting catastrophic forgetting when learning from non-stationary data streams. Compared to offline rehearsal, online rehearsal has faced particular challenges in tackling complex CL datasets due to the single-pass-through data constraint.  This work tries to analyze the internal workings of online rehearsal from a theoretical and conceptual perspective and identifies the fundamental challenge that it faces as the dilemma between overfitting locally and underfitting globally. To deal with this challenge, we propose a simple baseline: repeated and augmented rehearsal (RAR). Surprisingly, despite its simplicity, RAR achieves a large performance boost for a set of different rehearsal-based methods. 
Additionally, we propose an RL-based method to tune the hyperparameters of RAR to balance the stability-plasticity trade-off in an online manner. 
It achieves promising results compared to hyperparameter tuning based on validation data. 
This work is focused on continual learning for classification problems with image data. An interesting future research direction is to look at other CL domains, e.g., text/audio inputs or RL problems.

\medskip





\begin{ack}
This research is funded by the New Zealand MBIE
TAIAO data science programme.

\end{ack}
\bibliography{neurips_2022}

\begin{thebibliography}{24}
\providecommand{\natexlab}[1]{#1}
\providecommand{\url}[1]{\texttt{#1}}
\expandafter\ifx\csname urlstyle\endcsname\relax
  \providecommand{\doi}[1]{doi: #1}\else
  \providecommand{\doi}{doi: \begingroup \urlstyle{rm}\Url}\fi

\bibitem[Aljundi et~al.(2019)Aljundi, Belilovsky, Tuytelaars, Charlin, Caccia,
  Lin, and Page-Caccia]{aljundi2019online}
R.~Aljundi, E.~Belilovsky, T.~Tuytelaars, L.~Charlin, M.~Caccia, M.~Lin, and
  L.~Page-Caccia.
\newblock Online continual learning with maximal interfered retrieval.
\newblock \emph{Advances in Neural Information Processing Systems},
  32:\penalty0 11849--11860, 2019.

\bibitem[Bang et~al.(2021)Bang, Kim, Yoo, Ha, and Choi]{bang2021rainbow}
J.~Bang, H.~Kim, Y.~Yoo, J.-W. Ha, and J.~Choi.
\newblock Rainbow memory: Continual learning with a memory of diverse samples.
\newblock In \emph{Proceedings of the IEEE/CVF Conference on Computer Vision
  and Pattern Recognition}, pages 8218--8227, 2021.

\bibitem[Buzzega et~al.(2020)Buzzega, Boschini, Porrello, Abati, and
  Calderara]{buzzega2020dark}
P.~Buzzega, M.~Boschini, A.~Porrello, D.~Abati, and S.~Calderara.
\newblock Dark experience for general continual learning: a strong, simple
  baseline.
\newblock \emph{Advances in neural information processing systems},
  33:\penalty0 15920--15930, 2020.

\bibitem[Cha et~al.(2021)Cha, Lee, and Shin]{cha2021co2l}
H.~Cha, J.~Lee, and J.~Shin.
\newblock \text{Co2L}: Contrastive continual learning.
\newblock In \emph{Proceedings of the IEEE/CVF International Conference on
  Computer Vision}, pages 9516--9525, 2021.

\bibitem[Chaudhry et~al.(2018)Chaudhry, Ranzato, Rohrbach, and
  Elhoseiny]{chaudhry2018efficient}
A.~Chaudhry, M.~Ranzato, M.~Rohrbach, and M.~Elhoseiny.
\newblock Efficient lifelong learning with a-gem.
\newblock In \emph{International Conference on Learning Representations}, 2018.

\bibitem[Chaudhry et~al.(2019)Chaudhry, Rohrbach, Elhoseiny, Ajanthan, Dokania,
  Torr, and Ranzato]{chaudhry2019tiny}
A.~Chaudhry, M.~Rohrbach, M.~Elhoseiny, T.~Ajanthan, P.~K. Dokania, P.~H. Torr,
  and M.~Ranzato.
\newblock On tiny episodic memories in continual learning.
\newblock \emph{arXiv preprint arXiv:1902.10486}, 2019.

\bibitem[Chen et~al.(2020)Chen, Dobriban, and Lee]{chen2020group}
S.~Chen, E.~Dobriban, and J.~H. Lee.
\newblock A group-theoretic framework for data augmentation.
\newblock \emph{Journal of Machine Learning Research}, 21\penalty0
  (245):\penalty0 1--71, 2020.

\bibitem[Cubuk et~al.(2020)Cubuk, Zoph, Shlens, and Le]{cubuk2020randaugment}
E.~D. Cubuk, B.~Zoph, J.~Shlens, and Q.~V. Le.
\newblock Randaugment: Practical automated data augmentation with a reduced
  search space.
\newblock In \emph{Proceedings of the IEEE/CVF Conference on Computer Vision
  and Pattern Recognition Workshops}, pages 702--703, 2020.

\bibitem[Delange et~al.(2021)Delange, Aljundi, Masana, Parisot, Jia, Leonardis,
  Slabaugh, and Tuytelaars]{delange2021continual}
M.~Delange, R.~Aljundi, M.~Masana, S.~Parisot, X.~Jia, A.~Leonardis,
  G.~Slabaugh, and T.~Tuytelaars.
\newblock A continual learning survey: Defying forgetting in classification
  tasks.
\newblock \emph{IEEE Transactions on Pattern Analysis and Machine
  Intelligence}, 2021.

\bibitem[Grossberg(2012)]{grossberg2012studies}
S.~T. Grossberg.
\newblock \emph{Studies of mind and brain: Neural principles of learning,
  perception, development, cognition, and motor control}, volume~70.
\newblock Springer Science \& Business Media, 2012.

\bibitem[Li et~al.(2020)Li, Jiang, Gu, Peng, Li, Hong, and Tao]{li2020clrs}
H.~Li, H.~Jiang, X.~Gu, J.~Peng, W.~Li, L.~Hong, and C.~Tao.
\newblock \text{CLRS}: Continual learning benchmark for remote sensing image
  scene classification.
\newblock \emph{Sensors}, 20\penalty0 (4):\penalty0 1226, 2020.

\bibitem[Li and Hoiem(2017)]{li2017learning}
Z.~Li and D.~Hoiem.
\newblock Learning without forgetting.
\newblock \emph{IEEE Transactions on Pattern Analysis and Machine
  Intelligence}, 40\penalty0 (12):\penalty0 2935--2947, 2017.

\bibitem[Lomonaco and Maltoni(2017)]{lomonaco2017core50}
V.~Lomonaco and D.~Maltoni.
\newblock \text{CORe50}: a new dataset and benchmark for continuous object
  recognition.
\newblock In \emph{Conference on Robot Learning}, pages 17--26. PMLR, 2017.

\bibitem[Lopez-Paz and Ranzato(2017)]{lopez2017gradient}
D.~Lopez-Paz and M.~Ranzato.
\newblock Gradient episodic memory for continual learning.
\newblock \emph{Advances in neural information processing systems}, 30, 2017.

\bibitem[Mai et~al.(2021)Mai, Li, Kim, and Sanner]{mai2021supervised}
Z.~Mai, R.~Li, H.~Kim, and S.~Sanner.
\newblock Supervised contrastive replay: Revisiting the nearest class mean
  classifier in online class-incremental continual learning.
\newblock In \emph{Proceedings of the IEEE/CVF Conference on Computer Vision
  and Pattern Recognition}, pages 3589--3599, 2021.

\bibitem[Mai et~al.(2022)Mai, Li, Jeong, Quispe, Kim, and
  Sanner]{mai2022online}
Z.~Mai, R.~Li, J.~Jeong, D.~Quispe, H.~Kim, and S.~Sanner.
\newblock Online continual learning in image classification: An empirical
  survey.
\newblock \emph{Neurocomputing}, 469:\penalty0 28--51, 2022.

\bibitem[Mirzadeh et~al.(2020)Mirzadeh, Farajtabar, Gorur, Pascanu, and
  Ghasemzadeh]{mirzadeh2020linear}
S.~I. Mirzadeh, M.~Farajtabar, D.~Gorur, R.~Pascanu, and H.~Ghasemzadeh.
\newblock Linear mode connectivity in multitask and continual learning.
\newblock In \emph{International Conference on Learning Representations}, 2020.

\bibitem[Rebuffi et~al.(2017)Rebuffi, Kolesnikov, Sperl, and
  Lampert]{rebuffi2017icarl}
S.-A. Rebuffi, A.~Kolesnikov, G.~Sperl, and C.~H. Lampert.
\newblock \text{iCaRL}: Incremental classifier and representation learning.
\newblock In \emph{Proceedings of the IEEE conference on Computer Vision and
  Pattern Recognition}, pages 2001--2010, 2017.

\bibitem[Shim et~al.(2021)Shim, Mai, Jeong, Sanner, Kim, and
  Jang]{shim2021online}
D.~Shim, Z.~Mai, J.~Jeong, S.~Sanner, H.~Kim, and J.~Jang.
\newblock Online class-incremental continual learning with adversarial shapley
  value.
\newblock In \emph{Proceedings of the AAAI Conference on Artificial
  Intelligence}, volume~35, pages 9630--9638, 2021.

\bibitem[Verwimp et~al.(2021)Verwimp, De~Lange, and
  Tuytelaars]{verwimp2021rehearsal}
E.~Verwimp, M.~De~Lange, and T.~Tuytelaars.
\newblock Rehearsal revealed: The limits and merits of revisiting samples in
  continual learning.
\newblock In \emph{Proceedings of the IEEE/CVF International Conference on
  Computer Vision (ICCV)}, pages 9385--9394, 2021.

\bibitem[Vinyals et~al.(2016)Vinyals, Blundell, Lillicrap, and
  Wierstra]{vinyals2016matching}
O.~Vinyals, C.~Blundell, T.~Lillicrap, and D.~Wierstra.
\newblock Matching networks for one shot learning.
\newblock \emph{Advances in neural information processing systems},
  29:\penalty0 3630--3638, 2016.

\bibitem[Vitter(1985)]{vitter1985random}
J.~S. Vitter.
\newblock Random sampling with a reservoir.
\newblock \emph{ACM Transactions on Mathematical Software (TOMS)}, 11\penalty0
  (1):\penalty0 37--57, 1985.

\bibitem[Wu et~al.(2019)Wu, Chen, Wang, Ye, Liu, Guo, and Fu]{wu2019large}
Y.~Wu, Y.~Chen, L.~Wang, Y.~Ye, Z.~Liu, Y.~Guo, and Y.~Fu.
\newblock Large scale incremental learning.
\newblock In \emph{Proceedings of the IEEE/CVF Conference on Computer Vision
  and Pattern Recognition}, pages 374--382, 2019.

\bibitem[Zhang and Goh(2019)]{zhang2019bootstrapped}
Y.~Zhang and W.-B. Goh.
\newblock Bootstrapped policy gradient for difficulty adaptation in intelligent
  tutoring systems.
\newblock In \emph{Proceedings of the 18th International Conference on
  Autonomous Agents and MultiAgent Systems}, pages 711--719, 2019.

\end{thebibliography}
\bibliographystyle{abbrvnat}
\section*{Checklist}

The checklist follows the references.  Please
read the checklist guidelines carefully for information on how to answer these
questions.  For each question, change the default \answerTODO{} to \answerYes{},
\answerNo{}, or \answerNA{}.  You are strongly encouraged to include a {\bf
justification to your answer}, either by referencing the appropriate section of
your paper or providing a brief inline description.  For example:
\begin{itemize}
  \item Did you include the license to the code and datasets? \answerYes{}
\end{itemize}
Please do not modify the questions and only use the provided macros for your
answers.  Note that the Checklist section does not count towards the page
limit.  In your paper, please delete this instructions block and only keep the
Checklist section heading above along with the questions/answers below.

\begin{enumerate}

\item For all authors...
\begin{enumerate}
  \item Do the main claims made in the abstract and introduction accurately reflect the paper's contributions and scope?
  \answerYes{}
  \item Did you describe the limitations of your work?
    \answerYes{}
  \item Did you discuss any potential negative societal impacts of your work?
     \answerNA{}
  \item Have you read the ethics review guidelines and ensured that your paper conforms to them?
  \answerYes{}
\end{enumerate}

\item If you are including theoretical results...
\begin{enumerate}
  \item Did you state the full set of assumptions of all theoretical results?
     \answerYes{}
        \item Did you include complete proofs of all theoretical results?
         \answerYes{}
\end{enumerate}

\item If you ran experiments...
\begin{enumerate}
  \item Did you include the code, data, and instructions needed to reproduce the main experimental results (either in the supplemental material or as a URL)?
     \answerYes{}
  \item Did you specify all the training details (e.g., data splits, hyperparameters, how they were chosen)?
     \answerYes{}
        \item Did you report error bars (e.g., with respect to the random seed after running experiments multiple times)?
     \answerYes{}
        \item Did you include the total amount of compute and the type of resources used (e.g., type of GPUs, internal cluster, or cloud provider)?
     \answerYes{}
\end{enumerate}

\item If you are using existing assets (e.g., code, data, models) or curating/releasing new assets...
\begin{enumerate}
  \item If your work uses existing assets, did you cite the creators?
     \answerYes{}
  \item Did you mention the license of the assets?
     \answerYes{}
  \item Did you include any new assets either in the supplemental material or as a URL?
     \answerNA{}
  \item Did you discuss whether and how consent was obtained from people whose data you're using/curating?
    \answerNA{}
  \item Did you discuss whether the data you are using/curating contains personally identifiable information or offensive content?
    \answerNA{}
\end{enumerate}

\item If you used crowdsourcing or conducted research with human subjects...
\begin{enumerate}
  \item Did you include the full text of instructions given to participants and screenshots, if applicable?
    \answerNA{}
  \item Did you describe any potential participant risks, with links to Institutional Review Board (IRB) approvals, if applicable?
    \answerNA{}
  \item Did you include the estimated hourly wage paid to participants and the total amount spent on participant compensation?
    \answerNA{}
\end{enumerate}

\end{enumerate}

\newpage
\appendix
\onecolumn
\section{Theoretical Analysis}
\label{sec:theory}
This section contains the theoretical analysis of the loss functions of offline experience replay (\textbf{Proposition 2}), augmented experience replay (\textbf{Proposition 3}), and online experience replay with reservoir sampling (\textbf{Proposition 1}).

\textbf{Proposition 2 (Empirical risk minimization for experience replay)}: We assume a memory set $\mathcal{D}_\mathcal{M}$ and an incoming task stream $\mathcal{D}_\mathcal{T}$ with different data distribution $\mathbb{P}(\mathcal{D}_\mathcal{T})\neq \mathbb{P}(\mathcal{D}_\mathcal{M})$. At each iteration $t$, a batch of data is sampled from memory $\mathcal{B}^\mathcal{M}_t\sim \mathcal{D}_\mathcal{M}$ with $\mathcal{B}^\mathcal{M}_t=\{\textbf{x}_i,y_i\}_{i=1,...|\mathcal{B}^\mathcal{M}_t|}$, and a batch of data is sampled from incoming task $\mathcal{B}_t\sim \mathcal{D}_\mathcal{T}$ with $\mathcal{B}_t=\{\textbf{x}_i,y_i\}_{i=1,...|\mathcal{B}_t|}$. To update the parameter $\theta$ of a function  $f_\theta$:  $f_\theta \times \mathcal{X}\rightarrow \mathcal{Y}$ based on loss function $\mathcal{L}$, consider the parameter update rule defined by  $\theta=\theta- \frac{\eta}{\left|\mathcal{B}_{t}\right|} \sum_{\textbf{x}_i,y_i \in  \mathcal{B}_t} \nabla \mathcal{L}\left(f_\theta(\textbf{x}_{i}),y_i\right) -\frac{\eta}{\left|\mathcal{B}^\mathcal{M}_{t}\right|} \sum_{\textbf{x}_i,y_i \in \mathcal{B}^\mathcal{M}_{t}} \nabla \mathcal{L}\left(f_\theta(\textbf{x}_{i}),y_i\right) $, then it is an unbiased stochastic gradient descent update rule for the following empirical risk: $$ \mathcal{R}(\theta)= \sum_{x_i,y_i\in \mathcal{D}_\mathcal{T}} \mathcal{L}(f_\theta({\textbf{x}}_i),y_i)+\frac{|\mathcal{D}_\mathcal{T}|}{|\mathcal{D}_\mathcal{M}|}\sum_{x_i,y_i \in \mathcal{D}_\mathcal{M}}\mathcal{L}(f_\theta(\textbf{x}_i),y_i).$$

\textbf{Proof}:
Given the empirical gradient $\hat{g}_{ER}= \frac{1}{\left|\mathcal{B}_{t}\right|} \sum_{\textbf{x}_i,y_i \in  \mathcal{B}_t} \nabla \mathcal{L}\left(f_\theta(\textbf{x}_{i}),y_i\right) +\frac{1}{\left|\mathcal{B}^\mathcal{M}_{t}\right|} \sum_{\textbf{x}_i,y_i \in \mathcal{B}^\mathcal{M}_{t}} \nabla \mathcal{L}\left(f_\theta(\textbf{x}_{i}),y_i\right)
  $ during stochastic optimization, we can derive the gradient expectation as follows: 
\begin{equation*}
\begin{aligned}
  \mathbb{E}_{\mathcal{B}^\mathcal{M}_t\sim \mathcal{D}_\mathcal{M},\mathcal{B}_t\sim \mathcal{D}_\mathcal{T}} [\hat{g}_{ER}]&= \nabla (  \mathbb{E}_{ {\mathcal{B}^\mathcal{M}_t\sim\mathcal{D}_\mathcal{M}}}[\frac{1}{\left|\mathcal{B}^\mathcal{M}_t\right|}\sum_{\mathcal{B}^\mathcal{M}_t} \mathcal{L}\left(f_\theta( \textbf{x}_{i}),y_i\right)]+ \mathbb{E}_{\mathcal{B}_t \sim {\mathcal{D}_\mathcal{T}}}[  \frac{1}{\left|\mathcal{B}_t\right|}\sum_{\mathcal{B}_t} \mathcal{L}\left(f_\theta( \textbf{x}_{i}),y_i\right)])\\
  &= \nabla ( \mathbb{E}_{ {x_i,y_i \sim \mathcal{D}_\mathcal{M}}}[ \mathcal{L}\left(f_\theta( \textbf{x}_{i}),y_i\right)]+ \mathbb{E}_{ x_i,y_i\sim {\mathcal{D}_\mathcal{T}}}[ \mathcal{L}\left(f_\theta( \textbf{x}_{i}),y_i\right)] )\\
  &= \nabla ( \frac{1 }{|\mathcal{D}^\mathcal{M}|}\sum_{ {\mathcal{D}_\mathcal{M}}} \mathcal{L}\left(f_\theta( \textbf{x}_{i}),y_i\right)+\frac{1 }{|\mathcal{D}_\mathcal{T}|} \sum_{ {\mathcal{D}_\mathcal{T}}} \mathcal{L}\left(f_\theta( \textbf{x}_{i}),y_i\right) )\\
   & =\frac{1}{|\mathcal{D}_\mathcal{T}|} \nabla \left( \frac{|\mathcal{D}_\mathcal{T}|}{|\mathcal{D}_\mathcal{M}|} \sum_{ {\mathcal{D}_\mathcal{M}}} \mathcal{L}\left(f_\theta( \textbf{x}_{i}),y_i\right)+ \sum_{ {\mathcal{D}_\mathcal{T}}} \mathcal{L}\left(f_\theta( \textbf{x}_{i}),y_i\right)\right)\\
  \end{aligned}
\end{equation*}

\textbf{Proposition 3 (Augmented empirical risk minimization for experience  replay)}: Assume a memory set $\mathcal{D}_\mathcal{M}=\{\textbf{x}_i,y_i\}_{i=1,..,|\mathcal{D}_\mathcal{M}|}$ and an incoming task stream $\mathcal{D}_\mathcal{T}=\{\textbf{x}_i,y_i\}_{i=1,..,|\mathcal{D}_\mathcal{T}|}$ with different data distribution $\mathbb{P}(\mathcal{D}_\mathcal{T})\neq \mathbb{P}(\mathcal{D}_\mathcal{M})$, and a compact topological group of transform $G$ with a probability distribution $\mathbb{Q}$ that acts on the input space $\mathcal{X}$ and is invariant under function $f$, i.e., $f(gx)=f(x), g\in G, x \in \mathcal{X}$.  At each iteration $t$, a batch of data is sampled from memory, $\mathcal{B}^\mathcal{M}_t\sim \mathcal{D}_\mathcal{M}$ and $\mathcal{B}^\mathcal{M}_t=\{\textbf{x}_i,y_i\}_{i=1,...|\mathcal{B}^\mathcal{M}_t|}$, a batch of data is sampled from the incoming task, $\mathcal{B}_t\sim \mathcal{D}_\mathcal{T}$ and $\mathcal{B}_t=\{\textbf{x}_i,y_i\}_{i=1,...|\mathcal{B}_t|}$, and a group operation $g_t\sim G$ is randomly selected. To update the parameter $\theta$ of the function $f_\theta$ based on loss function $\mathcal{L}$, consider the parameter update rule defined by  $\theta=\theta- \frac{\eta}{\left|\mathcal{B}_{t}\right|} \sum_{\textbf{x}_i,y_i \in  \mathcal{B}_t} \nabla \mathcal{L}\left(f_\theta(g_t\textbf{x}_{i}),y_i\right) -\frac{\eta}{\left|\mathcal{B}^\mathcal{M}_{t}\right|} \sum_{\textbf{x}_i,y_i \in \mathcal{B}^\mathcal{M}_{t}} \nabla \mathcal{L}\left(f_\theta(g_t\textbf{x}_{i}),y_i\right)$, then it is an unbiased stochastic gradient descent update rule for the loss function $$ \bar{\mathcal{R}}(\theta)= \sum_{x_i,y_i\in \mathcal{D}_\mathcal{T}}\int_G\mathcal{L}(f_\theta(g{x}_i),y_i)d\mathbb{Q}(g)+\frac{|\mathcal{D}_\mathcal{T}|}{|\mathcal{D}_\mathcal{M}|}\sum_{x_i,y_i \in \mathcal{D}_\mathcal{M}} \int_G\mathcal{L}(f_\theta(gx_i),y_i)d\mathbb{Q}(g).$$

\textbf{Proof}:

Given the augmented empirical gradient
  during stochastic optimization, we can derive the gradient expectation as follows: 
\begin{equation*}
\begin{aligned}
  &\mathbb{E}_{\mathcal{B}^\mathcal{M}_t\sim \mathcal{D}_\mathcal{M},\mathcal{B}_t\sim \mathcal{D}_\mathcal{T}, g\sim \mathbb{Q}} [\hat{g}] \\
  &= \nabla(  \mathbb{E}_{ {\mathcal{B}^\mathcal{M}_t\sim\mathcal{D}_\mathcal{M}}}[\frac{1}{\left|\mathcal{B}^\mathcal{M}_t\right|}\sum_{\mathcal{B}^\mathcal{M}_t}\int_g \mathcal{L}\left(f_\theta( g\textbf{x}_{i}),y_i\right)d\mathbb{Q}(g)]+ \mathbb{E}_{\mathcal{B}_t \sim {\mathcal{D}_\mathcal{T}}}[  \frac{1}{\left|\mathcal{B}_t\right|}\sum_{\mathcal{B}_t}\int_g \mathcal{L}\left(f_\theta( \textbf{x}_{i}),y_i\right)d\mathbb{Q}(g)])\\
  &=\frac{1}{|\mathcal{D}_\mathcal{T}|} \nabla \left( \sum_{x_i,y_i\in \mathcal{D}_\mathcal{T}}\int_G\mathcal{L}(f_\theta(g{x}_i),y_i)d\mathbb{Q}(g)+\frac{|\mathcal{D}_\mathcal{T}|}{|\mathcal{D}_\mathcal{M}|}\sum_{x_i,y_i \in \mathcal{D}_\mathcal{M}} \int_G\mathcal{L}(f_\theta(gx_i),y_i)d\mathbb{Q}(g) \right)\\
  \end{aligned}
  \end{equation*}
The second equality is based on the results of Proposition 1, which we prove next.

\textbf{Proposition 1 (ERM for online rehearsal)}: Assume an initial memory set $\mathcal{D}^0_\mathcal{M}$ and an incoming task stream $\mathcal{D}_\mathcal{T}$ with different data distribution $\mathbb{P}(\mathcal{D}_\mathcal{T})\neq \mathbb{P}(\mathcal{D}_\mathcal{M})$. At each iteration $t$, $t=1,..T$, a batch of data is sampled from the incoming task, $\mathcal{B}_t\sim \mathcal{D}_\mathcal{T}$ and $\mathcal{B}_t=\{\textbf{x}_i,y_i\}_{i=1,...|\mathcal{B}_t|}$, and a batch of data is sampled from the memory, $\mathcal{B}^\mathcal{M}_t\sim \mathcal{D}^t_\mathcal{M}$ and $\mathcal{B}^\mathcal{M}_t=\{\textbf{x}_i,y_i\}_{i=1,...|\mathcal{B}^\mathcal{M}_t|}$.  To update the parameter $\theta$ of the function $f_\theta$:  $f_\theta \times \mathcal{X}\rightarrow \mathcal{Y}$ based on loss function $\mathcal{L}$, consider a parameter update rule defined by  $\theta=\theta- \frac{\eta}{\left|\mathcal{B}_{t}\right|} \sum_{\textbf{x}_i,y_i \in  \mathcal{B}_t} \nabla \mathcal{L}\left(f_\theta(\textbf{x}_{i}),y_i\right) -\frac{\eta}{\left|\mathcal{B}^\mathcal{M}_{t}\right|} \sum_{\textbf{x}_i,y_i \in \mathcal{B}^\mathcal{M}_{t}} \nabla \mathcal{L}\left(f_\theta(\textbf{x}_{i}),y_i\right) $. Assume the memory is updated at the end of each iteration using reservoir sampling~\cite{vitter1985random,chaudhry2019tiny} $\mathcal{D}^{t+1}_\mathcal{M}\leftarrow RS( \mathcal{D}^{t}_\mathcal{M}, \mathcal{B}_t$). Then the gradient update rule is an unbiased stochastic gradient descent update rule for the loss function $$ \mathcal{R}_t(\theta)= \sum_{x_i,y_i\in \mathcal{D}_\mathcal{T}} \mathcal{L}(f_\theta({x}_i),y_i)+\beta_t \frac{|\mathcal{D}_\mathcal{T}|}{|\mathcal{D}^{0}_\mathcal{M}|}\sum_{x_i,y_i \in \mathcal{D}^0_\mathcal{M}}\mathcal{L}(f_\theta(x_i),y_i)$$ 
where $\beta_t =\frac{1}{1+2*\frac{N^t_{cur}}{N_{past}}}$, $N^t_{cur} = \sum_{i=1}^{i=t}|\mathcal{B}_i|$ denotes the number of samples of the current task that have been seen so far and  $N_{past}=\sum_{j=1}^{j=\mathcal{T}}|\mathcal{D}_j|$ denotes the number of samples of past tasks.

\textbf{Note 1}: The objective function $\mathcal{R}_t(\theta)$ changes with respect to the batch number $t$ due to the changes in  $\beta_t$.

\textbf{Note 2}:$\frac{1}{1+2*\frac{|\mathcal{D}_\mathcal{T}|}{N_{past}}} \leq \beta_t \leq 1$  and $\beta_t$ is decreasing with the batch number $t$. At the start of a task, $\beta_{t=0}=\beta_{max}=1$. At the end of a task, $\beta_{t=T}=\beta_{min} = \frac{1}{1+2*\frac{|\mathcal{D}_\mathcal{T}|}{N_{past}}}$.

\textbf{Note 3}: Consider a balanced continual learning dataset (e.g., Split-CIFAR100, Split-Mini-ImageNet) where $|\mathcal{D}_j|=|\mathcal{D}_\mathcal{T}|,j=1,..\mathcal{T}$. Then, we have  $ \beta_{min} = \frac{\mathcal{T}-1}{\mathcal{T}+1}= (1-\frac{2}{\mathcal{T}+1}) $ and $\lim_{\mathcal{T} \rightarrow \infty}\beta_{min}=1$.

\textbf{Note 4}: Consider general continual learning datasets. As CL learns more tasks, $N_{past}$ increases, and  $\lim_{N_{past} \rightarrow \infty}\beta_{t}=\lim_{N_{past} \rightarrow \infty}\frac{1}{1+2*\frac{N^t_{cur}}{N_{past}}} = 1$.

\textbf{Proof}:
Given the empirical gradient $\hat{g}_{ER}= \frac{1}{\left|\mathcal{B}_{t}\right|} \sum_{\textbf{x}_i,y_i \in  \mathcal{B}_t} \nabla \mathcal{L}\left(f_\theta(\textbf{x}_{i}),y_i\right) +\frac{1}{\left|\mathcal{B}^\mathcal{M}_{t}\right|} \sum_{\textbf{x}_i,y_i \in \mathcal{B}^\mathcal{M}_{t}} \nabla \mathcal{L}\left(f_\theta(\textbf{x}_{i}),y_i\right)
  $ during stochastic optimization, we can derive the gradient expectation as follows: 
\begin{equation*}
\begin{aligned}
 &\mathbb{E}_{\mathcal{D}^t_\mathcal{M}}[ \mathbb{E}_{\mathcal{B}^\mathcal{M}_t\sim \mathcal{D}^t_\mathcal{M},\mathcal{B}_t\sim \mathcal{D}_\mathcal{T}} [\hat{g}_{ER}]]
 \\
 &= \nabla  \mathbb{E}_{\mathcal{D}^t_\mathcal{M}}\left[  \mathbb{E}_{ {\mathcal{B}^\mathcal{M}_t\sim\mathcal{D}^t_\mathcal{M}}}[\frac{1}{\left|\mathcal{B}^\mathcal{M}_t\right|}\sum_{\mathcal{B}^\mathcal{M}_t} \mathcal{L}\left(f_\theta( \textbf{x}_{i}),y_i\right)]+ \mathbb{E}_{\mathcal{B}_t \sim {\mathcal{D}_\mathcal{T}}}[  \frac{1}{\left|\mathcal{B}_t\right|}\sum_{\mathcal{B}_t} \mathcal{L}\left(f_\theta( \textbf{x}_{i}),y_i\right)]\right]\\
  &= \nabla (  \mathbb{E}_{\mathcal{D}^t_\mathcal{M}}\left[\mathbb{E}_{ {x_i,y_i \sim \mathcal{D}^t_\mathcal{M}}}[ \mathcal{L}\left(f_\theta( \textbf{x}_{i}),y_i\right)]\right]+ \mathbb{E}_{ x_i,y_i\sim {\mathcal{D}_\mathcal{T}}}[ \mathcal{L}\left(f_\theta( \textbf{x}_{i}),y_i\right)] )\\
    &= \nabla (\frac{N_{past}}{N_{past}+N^t_{cur}}  \mathbb{E}_{ {x_i,y_i \sim \mathcal{D}^0_\mathcal{M}}}[ \mathcal{L}\left(f_\theta( \textbf{x}_{i}),y_i\right)]+(\frac{N^t_{cur}}{N_{past}+N^t_{cur}}+1) \mathbb{E}_{ x_i,y_i\sim {\mathcal{D}_\mathcal{T}}}[ \mathcal{L}\left(f_\theta( \textbf{x}_{i}),y_i\right)] )\\
  & =\frac{N_{past}+2*N^t_{cur}}{(N_{past}+N^t_{cur})*|\mathcal{D}_\mathcal{T}|} \nabla \left( \frac{N_{past}}{N_{past}+2*N^t_{cur}} \frac{|\mathcal{D}_\mathcal{T}|}{|\mathcal{D}^0_\mathcal{M}|} \sum_{ {\mathcal{D}^0_\mathcal{M}}} \mathcal{L}\left(f_\theta( \textbf{x}_{i}),y_i\right)+ \sum_{ {\mathcal{D}_\mathcal{T}}} \mathcal{L}\left(f_\theta( \textbf{x}_{i}),y_i\right)\right)\\
  \end{aligned}
\end{equation*}

The third equality is based on the condition that $\mathcal{D}^t_\mathcal{M}$ is updated using reservoir sampling so that all the data seen so far have an equal probability of being stored in the memory. In other words, at a given time $t$, the memory contains a fraction of  $\frac{N_{past}}{N_{past}+N^t_{cur}}$ samples coming from  past tasks and  a fraction of  $\frac{N^t_{curr}}{N_{past}+N^t_{cur}}$ samples coming from  the current task.

\section{Dataset details}
Table~\ref{tab:dataset} lists the image size, the number of classes, the number of tasks, and data size per task of the four CL benchmarks.
\label{sec:dataset}
\begin{table}[h]
\caption{Dataset information for the four CL benchmarks. } 
\label{tab:dataset}
\vskip 0.15in
\begin{center}
\begin{small}
\begin{sc}
\begin{tabular}{l|ccccc}
\toprule
 & Image Size & \#Task  &  \# Class  & Train per task & Test per task\\
\midrule
Seq-CIFAR100 & 3x32x32 &20 &100 &2,500 & 500\\
Seq-MINI-ImageNet &3x84x84 &10 &100 &5,000 & 1,000\\
CORE50-NC &3x128x128 &9 & 50 & 12,000-24,000 & 4,500-9,000\\
CLRS-NC & 3x256x256 & 5 & 25 & 2,250 & 750 \\
\bottomrule
\end{tabular}
\end{sc}
\end{small}
\end{center}
\vskip -0.1in
\end{table}
\section{Implementation Details}
\label{sec:hyper}
\subsection{Continual Learning Implementation}
The hyperparameter settings are summarized in Table~\ref{tab:hyper}. All models are optimized using vanilla SGD. For all experiments, we use
the learning rate of 0.1 following the same setting as in~\cite{aljundi2019online,shim2021online}, and the Nearest-Class-Mean (NCM) classifier is used for evaluation, as Mai et al. reported (\citeyear{mai2021supervised})
considerable and consistent performance gains when
replacing the Softmax classifier with the NCM classifier. Each mini batch
during training consists of 10 new and 10 memory samples,
except for the SCR method, which employs 100 memory samples and 10 new incoming samples~\cite{mai2021supervised}. By default, the repeated rehearsal parameter for all the results is $K=10$ and the augmented rehearsal parameters are $P=1, Q=14$.

This paper uses Randaugment~\citep{cubuk2020randaugment}, which is an auto augmentation method. It randomly selects $P$  augmentation operators from a set of 14 operators and applies them to the images. The augmentation operator set includes:’Identity’, ’AutoContrast’, ’Equalize’, ’Rotate’,
’Solarize’, ’Color’,’Posterize’, ’Contrast’, ’Brightness’, ’Sharpness’,’ShearX’, ’ShearY’,’TranslateX’, ’TranslateY’. 

\begin{table}[h]
\caption{Hyperparameter setting. } 
\label{tab:hyper}
\vskip 0.15in
\begin{center}
\begin{small}
\begin{sc}
\begin{tabular}{l|c}
\toprule
  & Hyperparameter\\
\midrule
LWF & lr=0.1\\
AGEM & lr=0.1\\
ER & batchsize=10,lr=0.1\\
MIR & batchsize=10,C=50,lr=0.1\\
ASER & k=3, n\_smp\_cls=1.5,batchsize=10,lr=0.1\\
SCR &  temp =0.07, batch size = 100,lr=0.1\\
DER & $\alpha=0.3$, augmentation: flip and crop, $K=50$\\
\bottomrule
\end{tabular}
\end{sc}
\end{small}
\end{center}
\vskip -0.1in
\end{table}
\subsection{RL-based hyperparameter tuning implementation}
\label{sec:rl_implementation}
The memory iteration choices are from 1 to 20 and the augmentation choices are $(1,5),(1,14),(2,14),(3,14),(4,14)$.
Action selection probabilities are modeled with softmax weight $\pi_{w}(a_i)=\frac{e^{w_{i}}}{\sum_k e^{w_k}}$. Bootstrapped policy gradient is used to adjust action weights:
$$g^{BPG}=\mathbb{E}_{a_{i} \sim \pi_{w}}\left[\left|r_{a_{i}}\right|\left(\nabla_{w} \log \widehat{\pi}_{w}^{+}\left(a_{i}\right)-\nabla_{w} \log \widehat{\pi}_{w}^{-}\left(a_{i}\right)\right)\right]$$
where $\hat{\pi}_{w}^{+}\left(a_{i}\right):=\Sigma_{a_{k} \in \mathcal{X}_{a_{i}}^{+}} \pi_{w}\left(a_{k}\right)$ and  $\hat{\pi}_{w}^{-}\left(a_{i}\right):=\Sigma_{a_{k} \in \mathcal{X}_{a_{i}}^{-}} \pi_{w}\left(a_{k}\right)$

\textbf{Better/Worse action set} The key of the BPG idea is to incorporate prior information into the construction of better and worse action sets. To apply BPG in the OCL environment, we propose to determine the  better/worse action set based on the feedback in the form of current memory batch accuracy $A_\mathcal{M}$, which reflects the memory overfitting level 
of the CL agent. We want the training memory accuracy to be neither too high nor too low. The desirable training memory accuracy is defined by $A^{*}_\mathcal{M}$.  We find a higher repeated replay iteration leads to higher memory accuracy. Therefore, the better/worse action set is defined as follows:
$$\mathcal{X}_{a}^{+}:= \begin{cases}\forall a_{k} \mid Iter\left(a_{k}\right)<Iter(a), & A_\mathcal{M}(a)>A^{*}_\mathcal{M} \\ 
\forall a_{k} \mid Iter\left(a_{k}\right)>Iter(a), & A_\mathcal{M}(a)<A^{*}_\mathcal{M} \\ 
\emptyset, & A_\mathcal{M}(a)=A^{*}_\mathcal{M}\end{cases}
\mathcal{X}_{a}^{-}:= \begin{cases}\forall a_{k} \mid Iter\left(a_{k}\right)>Iter(a), & A_\mathcal{M}(a)>A^{*}_\mathcal{M} \\ 
\forall a_{k} \mid Iter\left(a_{k}\right)<Iter(a), & A_\mathcal{M}(a)<A^{*}_\mathcal{M} \\ 
\forall a_{k} \mid Iter\left(a_{k}\right) \neq Iter(a), & A_\mathcal{M}(a)=A^{*}_\mathcal{M}\end{cases}$$
$$\mathcal{X}_{a}^{+}:= \begin{cases}\forall a_{k} \mid Aug\left(a_{k}\right)>Aug(a), & A_\mathcal{M}(a)>A^{*}_\mathcal{M} \\ 
\forall a_{k} \mid Aug\left(a_{k}\right)<Aug(a), & A_\mathcal{M}(a)<A^{*}_\mathcal{M} \\ 
\emptyset, & A_\mathcal{M}(a)=A^{*}_\mathcal{M}\end{cases}
\mathcal{X}_{a}^{-}:= \begin{cases}\forall a_{k} \mid Aug\left(a_{k}\right)<Aug(a), & A_\mathcal{M}(a)>A^{*}_\mathcal{M} \\ 
\forall a_{k} \mid Aug\left(a_{k}\right)>Aug(a), & A_\mathcal{M}(a)<A^{*}_\mathcal{M} \\ 
\forall a_{k} \mid Aug\left(a_{k}\right) \neq Aug(a), & A_\mathcal{M}(a)=A^{*}_\mathcal{M}\end{cases}$$
\textbf{Reward} A challenge in the hyperparameter tuning for the OCL setting is that the CL agent may face new tasks with unseen data distribution. Therefore, it is often infeasible to assume the existence of an external validation data containing all the tasks in advance for hyperparameter tuning. To achieve online hyperparameter tuning without external validation data, we propose to define the reward  based on the accuracy on the memory. Given a target memory accuracy $A^{*}_\mathcal{M}$, the reward is defined as $r=|A_\mathcal{M}-A^{*}_\mathcal{M}|$

\textbf{Non-stationarity} To address the non-stationary nature in the CL environment, we reset the weight of BPG to a uniform weight at the start of each task.

The analysis of the selected action can be found in Figures~\ref{fig:action_selection_dataset} and \ref{fig:action_selection} in Section~\ref{sec:action_selection}.

\begin{minipage}{0.6\textwidth}

\captionof{algorithm}{BPG-based RAR}\label{algo_RAR_BPG}
\raggedright
\hrule 
 $\mathcal{M}$ is the memory with fixed size,\\
 $ \mathcal{B}_t$ is the incoming batch from the current task, \\
 $\theta$ are the parameters of the CL network, \\
 $w$ are the parameters of the RL agent, \\
 $K$ is the number of memory iterations, \\
 $P,Q$ are the augmentation hyperparameters\\
 $A^*_{\mathcal{M}}$ target memory accuracy \\
\begin{algorithmic}[1]
 \Procedure{RAR}{$\mathcal{M}_t$,  $ \mathcal{B}_t$, $\theta_t$, $w_t$ }
    \State $K_t \sim \pi_{w^{iter}_t}$
     \State $P_t,Q_t \sim \pi_{w^{aug}_t}$
    \For{$k = 1,...,K_t$}
      \State  $\mathcal{B}^\mathcal{M}_{t,k} \sim MemRetrieval(\mathcal{M}_t)$
      \State $\mathcal{B}_{aug} \leftarrow {\color{black}aug(\mathcal{B}^\mathcal{M}_{t,k} \cup \mathcal{B}_t,P_t,Q_t)}$
     \State$ A_{\mathcal{M}}  \leftarrow MemAcc(\mathcal{M},\theta_{t,k})$
      \State 
      $\mathcal{X}^{+},\mathcal{X}^{-}\leftarrow ActionSet(A_{\mathcal{M}},A^{*}_{\mathcal{M}})$

      \State $\theta_{t,k+1} \leftarrow SGD(\mathcal{B}_{aug},
\theta_{t,k})  $
    \EndFor\label{euclidendwhile}
    \State $\mathcal{M}_{t+1} \leftarrow MemUpdate(\mathcal{M}_t,\mathcal{B}_t) $
  \State $w^{iter}_{t+1},w^{aug}_{t+1} \leftarrow UpdateRL(A_{\mathcal{M}},\mathcal{X}^{+},\mathcal{X}^{-})$
  \EndProcedure
  \hrule
\end{algorithmic}
\end{minipage}%
\section{Supplementary Experimental Results}

\subsection{Ablation studies for MIR-RAR, ASER-RAR and SCR-RAR}
\label{sec:RAR_SCR}
\begin{figure}
\centerline{\includegraphics[width=1\columnwidth]{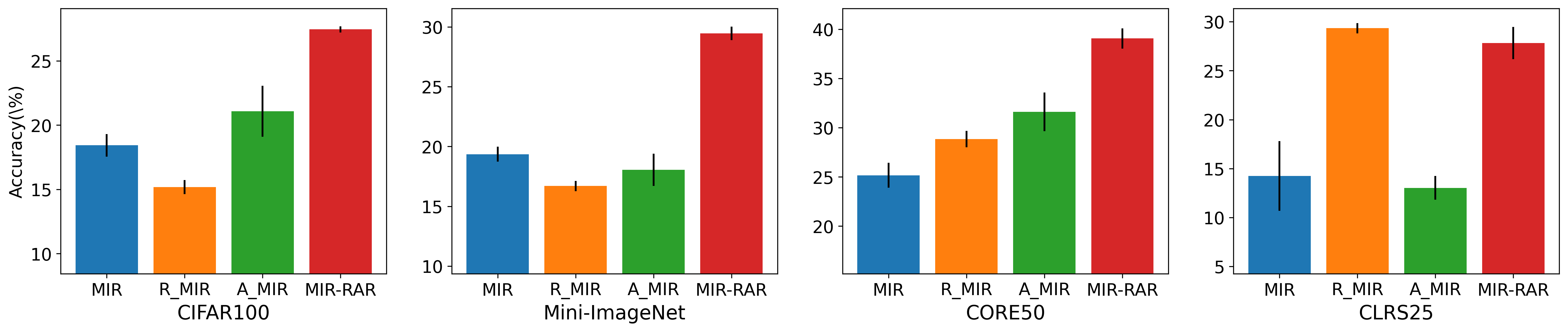}}
\caption{Ablation study: MIR-RAR on the four datasets, with a 2k memory}
\label{fig:mir_rar}
\end{figure}

\begin{figure}
\centerline{\includegraphics[width=1\columnwidth]{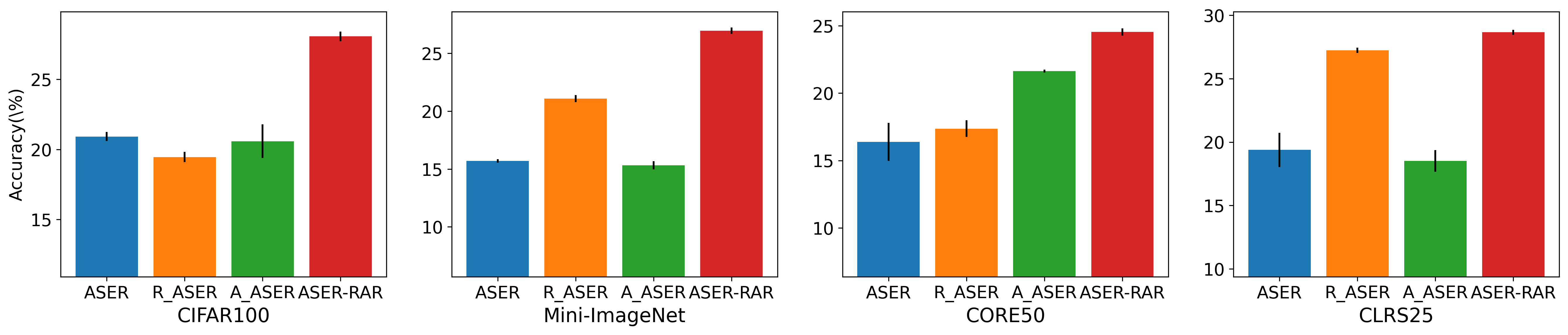}}
\caption{Ablation study: ASER-RAR on the four datasets, with a 2k memory}
\label{fig:aser_rar}
\end{figure}

\begin{figure}
    \centerline{\includegraphics[width=1\columnwidth]{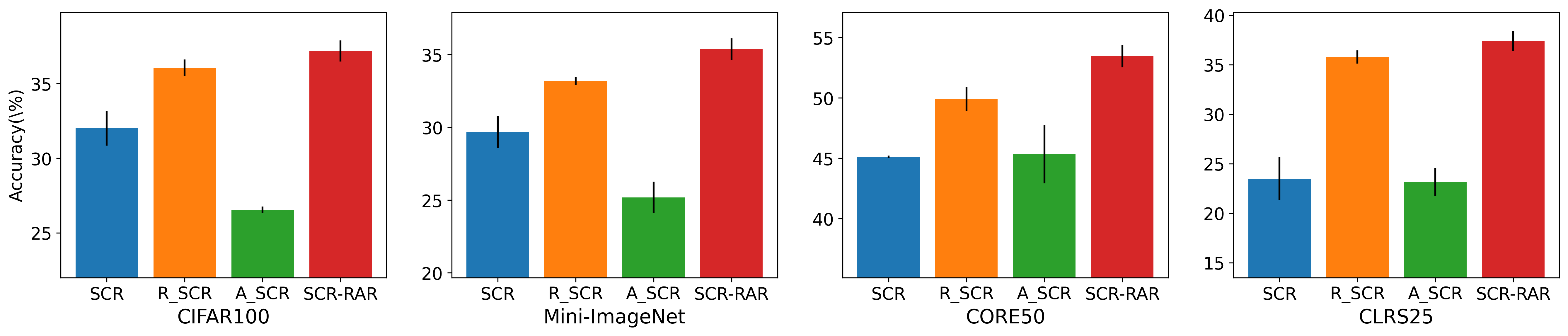}}
\caption{Ablation study: SCR-RAR on the four datasets, with a 2k memory}
\label{fig:rar_scr}
\end{figure}

As shown in Table~\ref{tab:main}, RAR improves ER and ER variants (MIR, ASER and SCR). The detailed results of the ablation studies of RAR with ER variants are presented in this section. In particular, Fig~\ref{fig:mir_rar} and \ref{fig:aser_rar} show the comparison of only using repeated rehearsal or augmented rehearsal for MIR and ASER respectively. Neither of them alone consistently improves the performance of the baseline. This result suggests the influence of RAR for MIR/ASER is similar to ER. The performance gain comes from the combination of the repeated rehearsal and augmented rehearsal. Fig~\ref{fig:rar_scr} shows the results for SCR. Repeated rehearsal  leads to consistent performance gains because SCR already includes augmentation.


\subsection{Reweighted ER}
\label{sec:reweighted_ER}
\begin{figure}
\centering
\subfigure[CIFAR100]{\includegraphics[width=0.4\columnwidth]{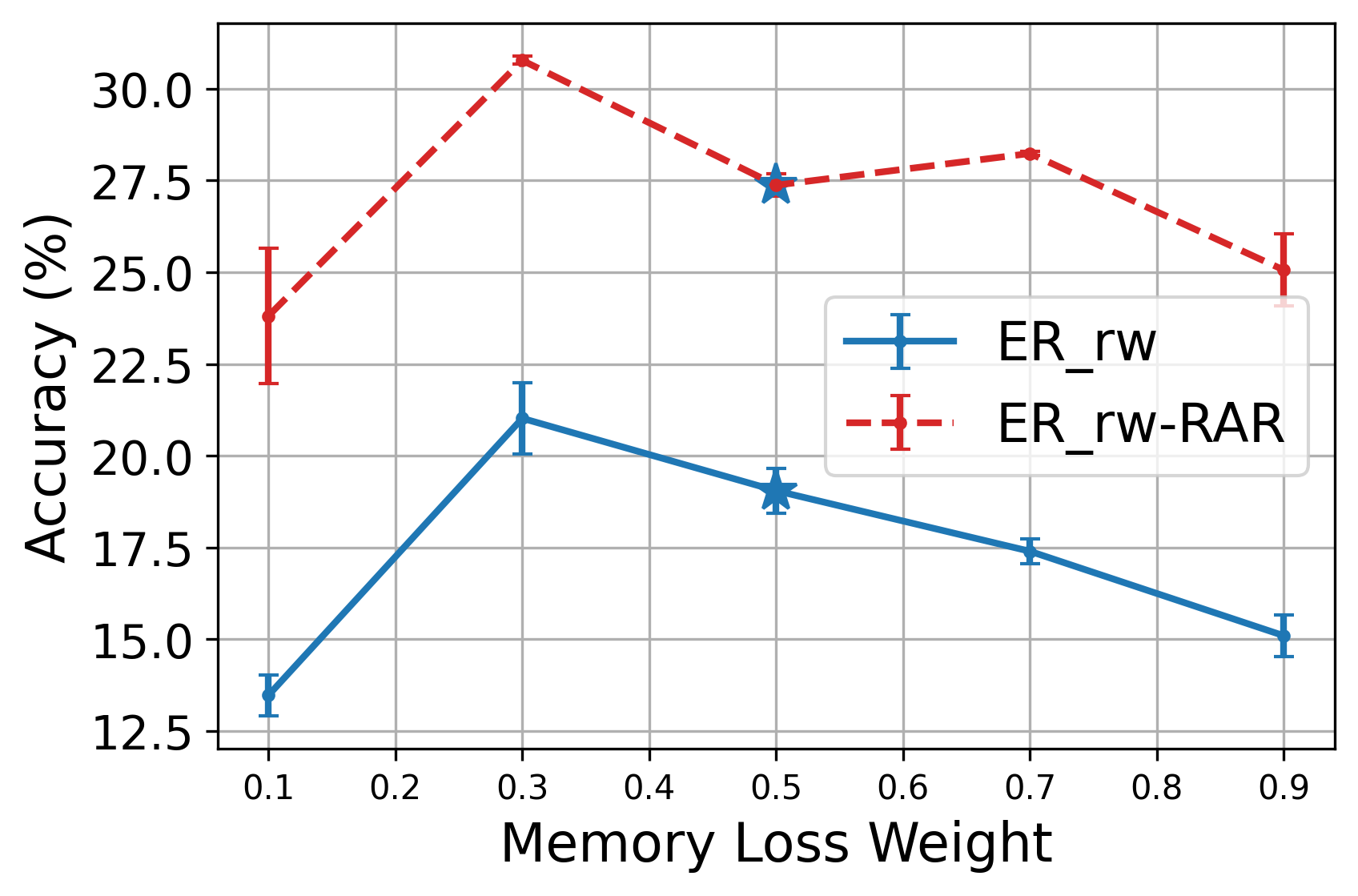}}
\subfigure[MINI-IMAGENET]{\includegraphics[width=0.4\columnwidth]{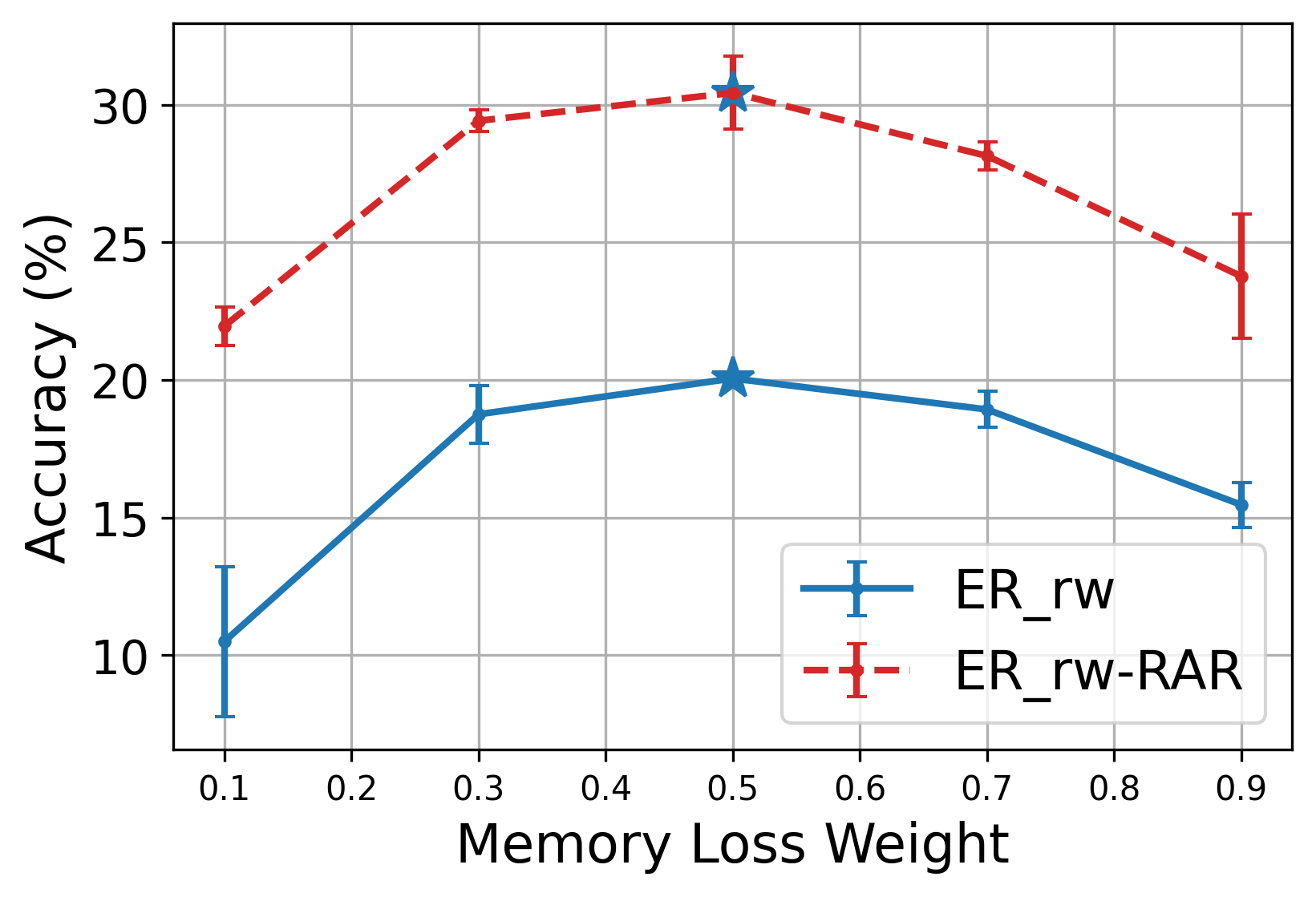}}
\subfigure[CORE50]{\includegraphics[width=0.4\columnwidth]{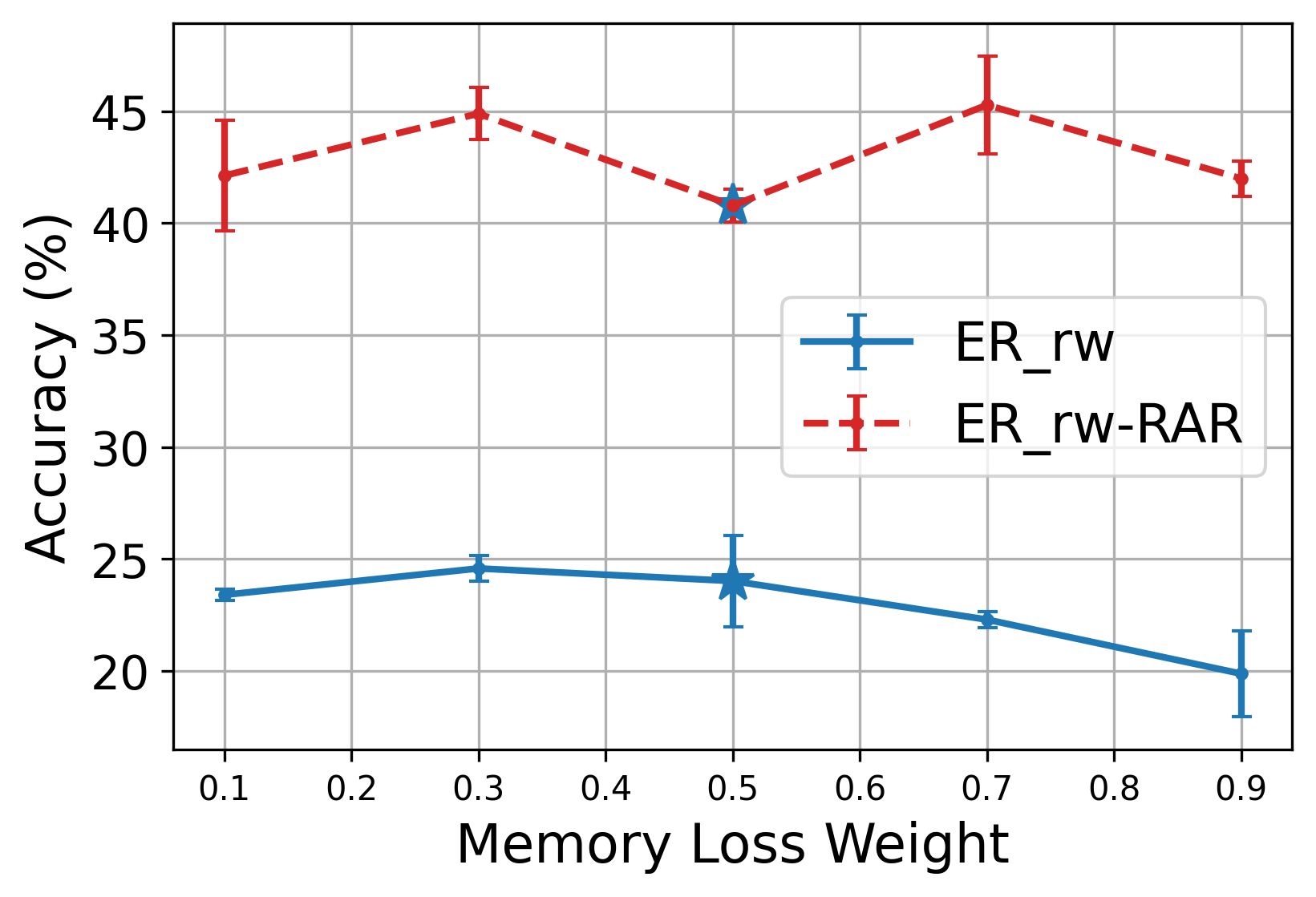}}
\subfigure[CLRS]{\includegraphics[width=0.4\columnwidth]{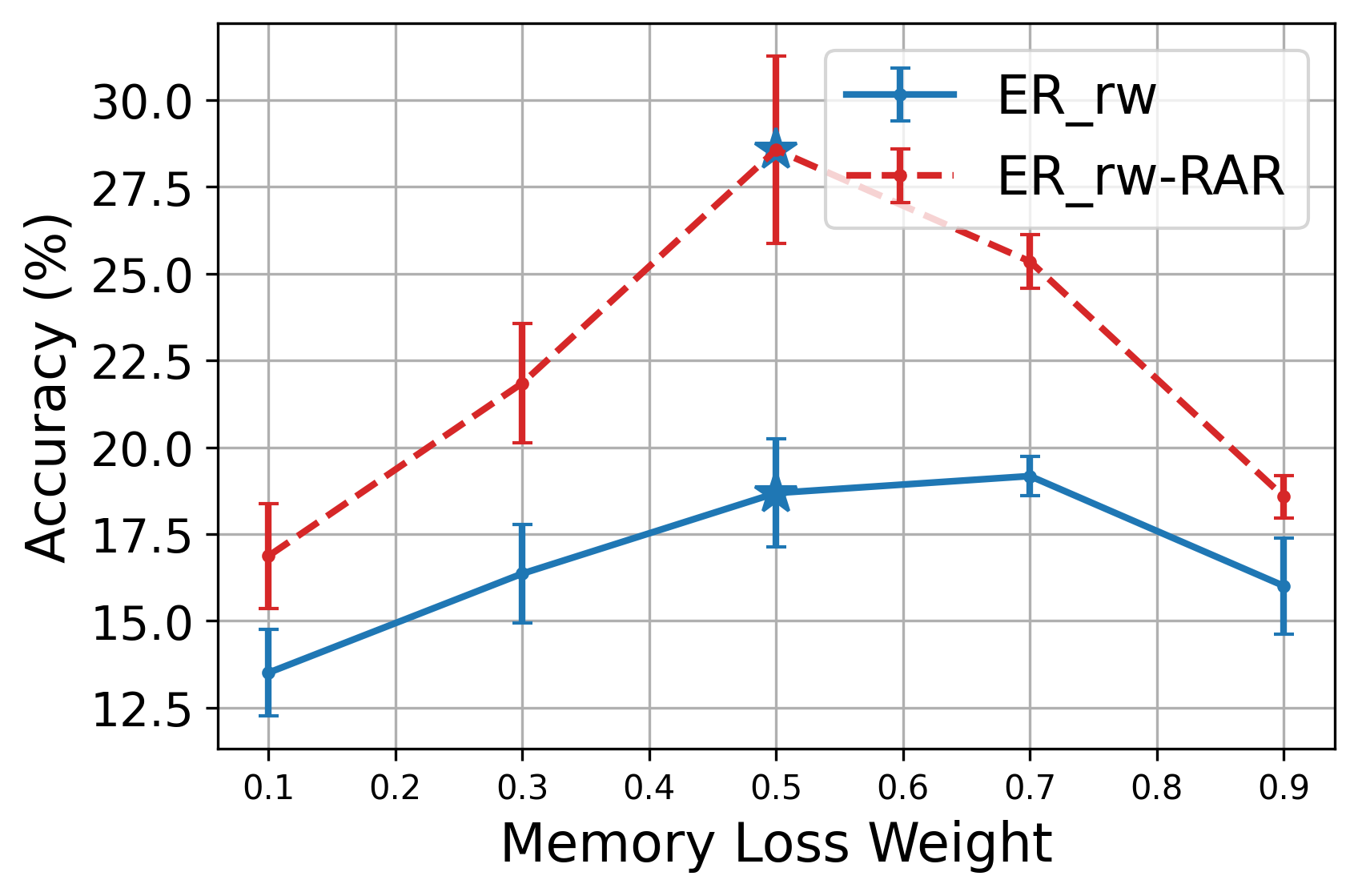}}

\caption{\color{black}The performance of reweighting the memory loss of ER (ER-rw) and its effectiveness with RAR on the four datasets. (Star symbols denote the accuracy of ER and ER-RAR with a reweighting value of 0.5).}
\label{fig:er-rw}
\end{figure}

{\color{black}
To deal with the biased ER loss, one straightforward way is to balance the weight of the memory loss and incoming loss by introducing a reweighting hyperparameter $\alpha$ in the gradient of Eq~\ref{eq:grad_er}. {\color{black} Specifically, the gradient for reweighted ER is implemented as 
$$g_{t}^{ER-rw}=(1-\alpha) \frac{1}{|B_t|} \sum_{x,y \in  B_t} \nabla{L}(f_\theta({x}),y) +\alpha\frac{1}{|B^M_t|} \sum_{x,y \in B^M_t} \nabla {L}(f_\theta({x}),y), \alpha \in (0,1).$$

The performance of ER and RAR with respect to  different reweighting hyperparameter values $\alpha \in [0.1,0.3,0.5,0.7,0.9]$ is shown in Fig~\ref{fig:er-rw}, where $\alpha=0.5$ denotes vanilla ER loss with the equal weighting of memory loss and incoming loss. To keep the learning rate comparable to vanilla ER, the learning rate of ER-rw is twice that of ER.} A key observation is that similar to vanilla ER, reweighted ER (ER-rw) also significantly benefits from repeated and augmented rehearsal, as ER-rw-rar greatly improves over ER-rw (see the red line and blue line in Fig~\ref{fig:er-rw})  for all four datasets. }


\subsection{Large-scale online CL}
\label{sec:imagenet}
{\color{black}

To examine the effectiveness of RAR in a large-scale continual learning problem, we apply RAR to ImageNet-1k. ImageNet-1k is split into 10 tasks and each task contains 100 classes.
The training dataset of 10 tasks contains 1,281,167 images in total. Considering the task size, we evaluate with a memory size of M= 20k (with the task to memory size ratio $\lambda\approx 6.4$) and a memory size of  M=100k (with the task to memory size ratio $\lambda\approx 1.28$). These choices are similar to Seq-CIFAR100 ($\lambda = 1.25$) and CORE50 ($\lambda=6$) with a 2k memory.
We randomly crop the images to size 224x224, and use the ResNet-18 architecture for training on ImageNet-1k with a single epoch. An incoming batch size of 32 and a memory batch size of 32 are used. 
We employ a learning rate of 0.1 with 0.001 of weight decay. For the RAR method, we use 10 memory iterations with RandAugment and parameters P=1 and Q=14. The results are shown in the
Table~\ref{tab:imagenet}. 

We observe that repeated augmented rehearsal (RAR) is effective in this large-scale online continual learning problem and improves the vanilla rehearsal from 2.1\% to 15.1\% with a 20k memory and from 8.7\% to 34.7\% with a 100k memory. 

The discussion in \cite{buzzega2020dark} suggests that to deal with complex continual learning datasets, it is necessary to employ multiple epochs of training with offline CL to avoid the underfitting problem present in online learning. Our results show that RAR can greatly improve online rehearsal for large-scale CL problems. Although several offline CL algorithms have been evaluated on ImageNet-1k, to our knowledge, our result is the first attempt to apply online CL to this problem. We aim to investigate other online rehearsal-based methods on ImageNet-1k in future work.

}

\begin{table}[h]
\caption{\color{black} Accuracy of ER and ER-RAR for ImageNet-1k  with a 20k and 100k memory. } 
\label{tab:imagenet}
\vskip 0.15in
\begin{center}
\begin{small}
\begin{sc}
\begin{tabular}{l|cc}
\toprule
 ImageNet-1k & M=20k & M=100k\\
\midrule
 ER        & 2.1 $\pm$ 0.3 & 8.7 $\pm$ 0.5  \\
 ER-RAR    & 15.1 $\pm$ 0.4 &34.7 $\pm$ 0.1  \\
Gains & 13.0 $\uparrow$ & 26.0 $\uparrow$ \\
\bottomrule
\end{tabular}
\end{sc}
\end{small}
\end{center}
\vskip -0.1in
\end{table}

\subsection{Action Selection in Reinforcement Learning}
\label{sec:action_selection}
\begin{figure}
    \centering
   \subfigure[RAR-RL]{ \includegraphics[width=0.4\columnwidth]{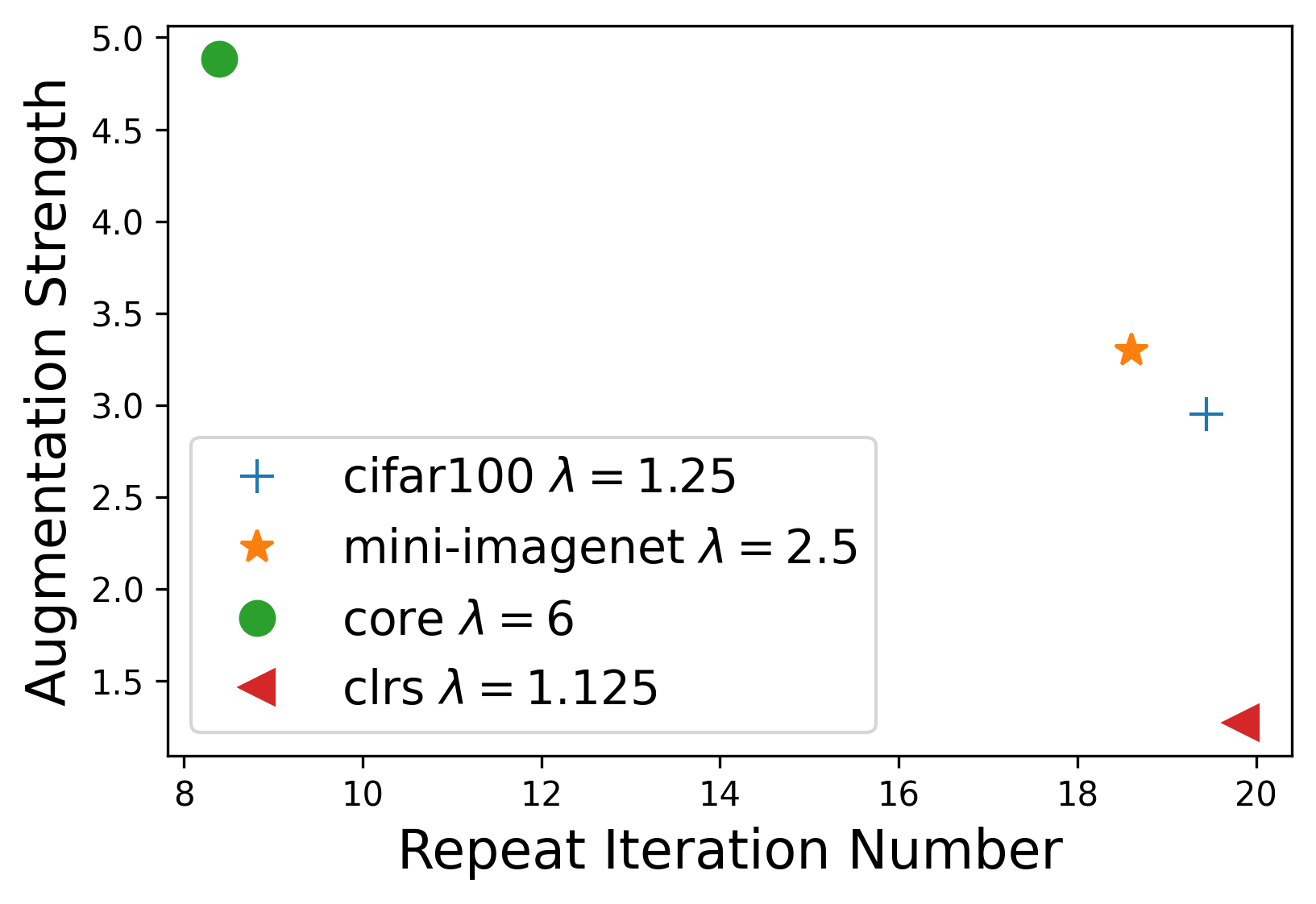}}
      \subfigure[RAR-HTOCL]{ \includegraphics[width=0.4\columnwidth]{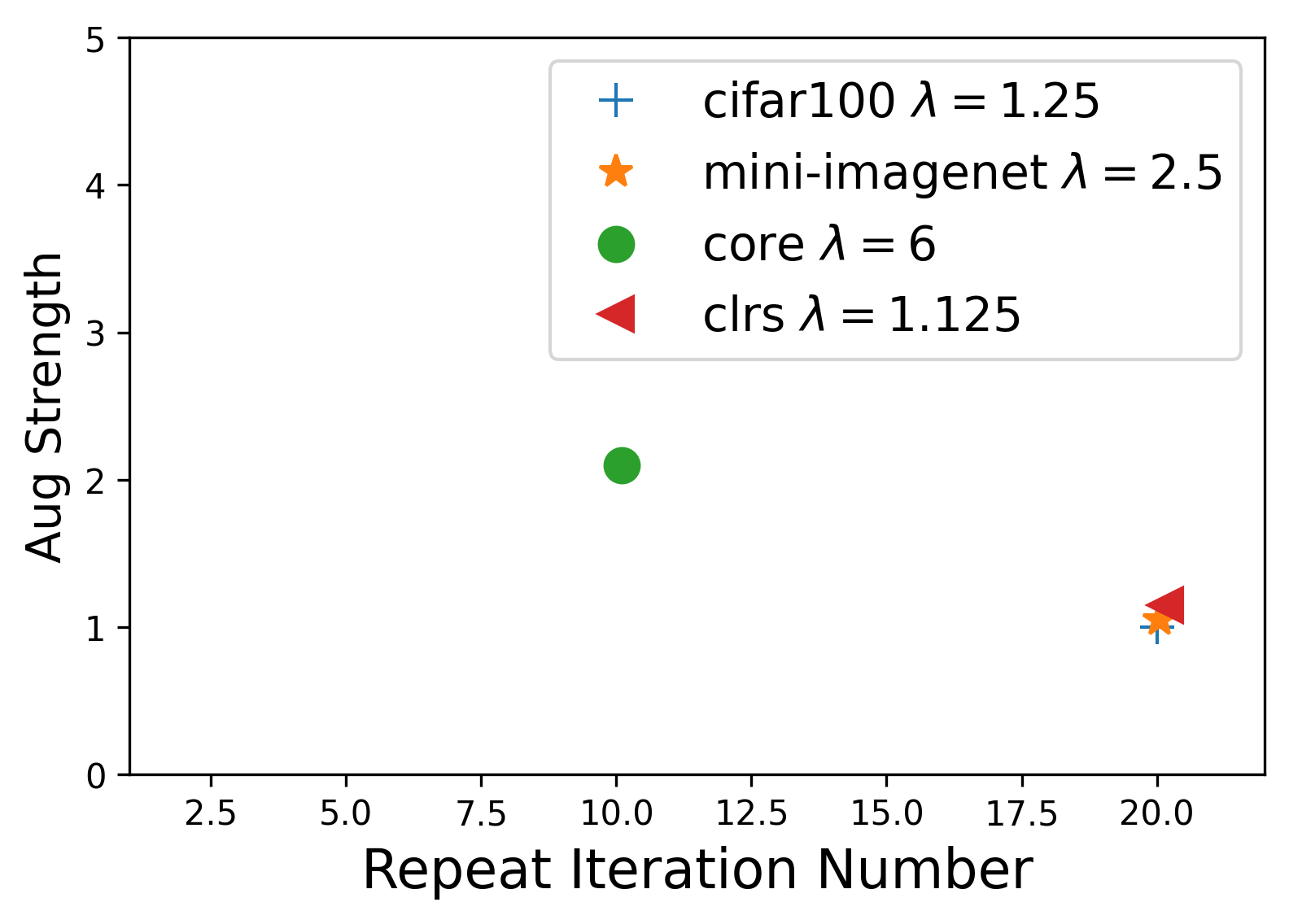}}
    \caption{The average of selected hyperparameters of RAR (iteration and augmentation values) for four datasets. }
    \label{fig:action_selection_dataset}
\end{figure}

The selected hyperparameters for four datasets are shown in Figure~\ref{fig:action_selection_dataset}. Interestingly,  the RL-based method assigns a stronger augmentation and lower iteration for a dataset with a higher $\lambda$ attribute. As discussed in the ERM analysis, a dataset with a higher $\lambda$ suggests a higher risk of overfitting. The RL-based method successfully takes this into account to tune the hyperparameters. In contrast, the OCL-HT method selects the weakest augmentation and highest iteration for three datasets.

\begin{figure}
    \centering
    \includegraphics[width=1\columnwidth]{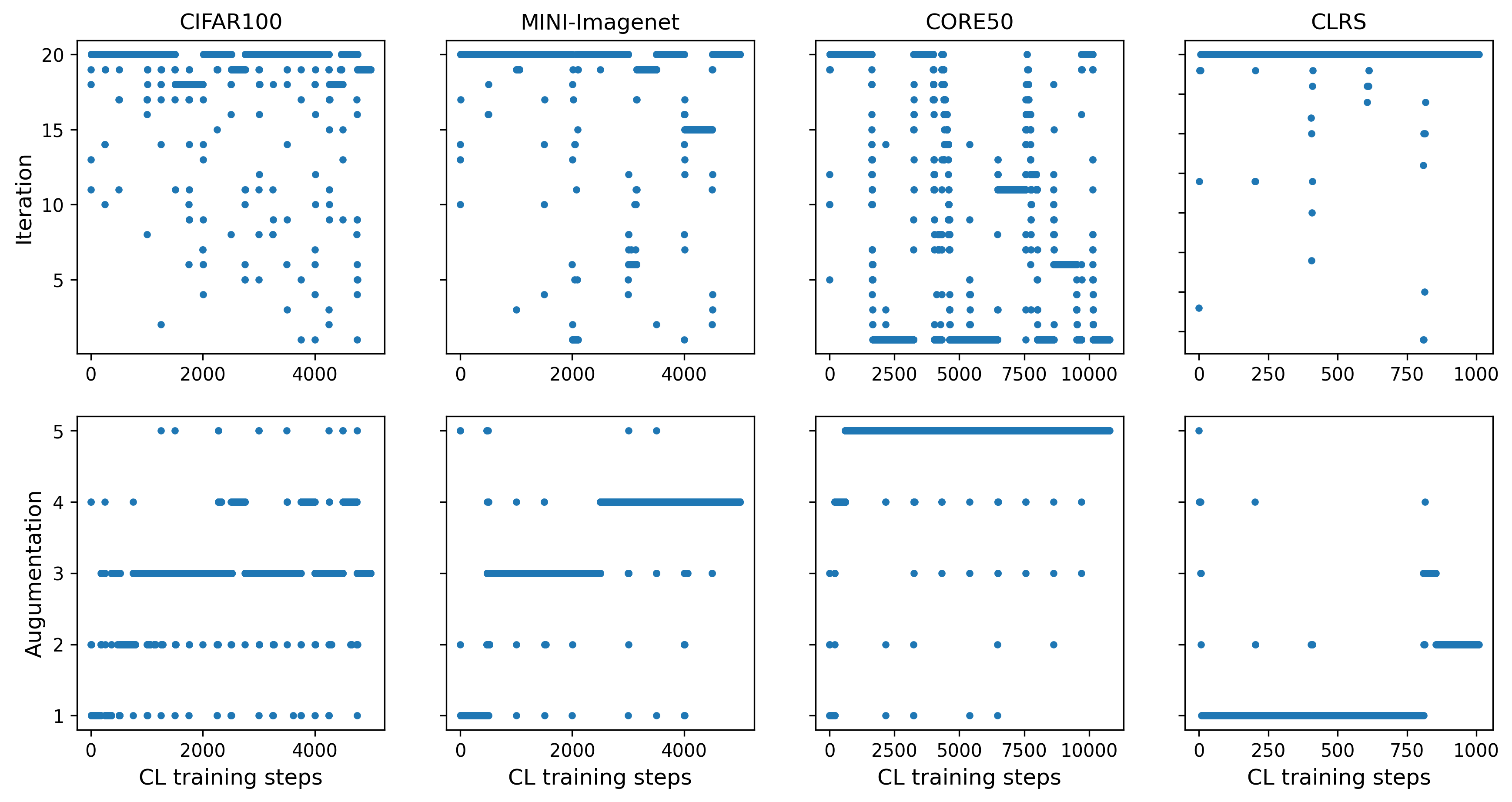}
    \caption{The selected hyperparameters of RAR (iteration and augmentation values) of the RL-based method. }
    \label{fig:action_selection}
\end{figure}

Figure~\ref{fig:action_selection} presents the selected hyperparameters at different CL training stages. Generally, as the continual learning proceeds, the RL-based method selects a stronger augmentation strength and lower iterations, to balance off the increasing risk of memory overfitting. 
\subsection{Running time}
\label{sec:running_time}
\subsubsection{Running time of RAR}
The running time of ER and ER-RAR is shown in Table~\ref{tab:running_time}. Experiments are conducted using an Nvidia GeForce RTX 2080 TI Graphics Card on Mini-ImageNet. The running time of RAR grows linearly with respect to the number of replay iterations. Hence,
compared to vanilla experience replay, RAR requires more running time due to the multiple iteration design. However,  RAR is still much more computationally efficient compared with offline CL with multiple epochs. An interesting future research direction is to study how to dynamically adjust the iteration number and augmentation strength of RAR to balance  the trade-off of accuracy and running time.

\begin{table}[h]
\caption{ Running time of ER and ER-RAR using an Nvidia GeForce RTX 2080 TI Graphics Card on Mini-ImageNet. The offline setting employs 50 epochs of training. } 
\label{tab:running_time}
\vskip 0.15in
\begin{center}
\begin{small}
\begin{sc}
\begin{tabular}{l|c|c}
\toprule
  & Running Time (s) & Accuracy (\%)\\
\midrule
ER  & 277 $\pm$	19 & 20.0 $\pm$	0.8 \\
RAR  ($K=5$) &1383 $\pm$ 4  &  29.1 $\pm$	0.9\\
RAR  ($K=10$) & 2,345 $\pm$ 4 & 30.4 $\pm$	1.3\\
R-ER ($K=10$) & 2,236 $\pm$ 1 & 17.8 $\pm$	0.6\\
\hline 
ER-offline ($E=50$)  &10,878 $\pm$ 31 & 20.4 $\pm$	0.6\\
RAR-RL ($K=19.6$) & 18,037 $\pm$ 71 & 32.1 $\pm$ 1.0\\
\bottomrule
\end{tabular}
\end{sc}
\end{small}
\end{center}
\vskip -0.1in
\end{table}

\subsubsection{Running time of RL}
{\color{black}
RL-based hyperparameter optimization is an online method that does not require repeated running over different hyperparameter choices. Therefore, RL-based HPO is much more computationally efficient than offline hyperparameter selection methods. More specifically, the hyperparameter search space in our problem is 100 with 5 augmentation strength levels and 20 memory iteration numbers. Grid search would need to run the OCL algorithm 100 times while RL only needs to run it once.

Nevertheless, the training of RL agents indeed introduces extra computation. In practice, we observe the running time of RL-RAR is about two times slower than that of RAR, as shown in the Table~\ref{tab:rl_running_time}. }
\begin{table}[h]
\caption{\color{black} Running time with and without RL-based hyperparameter optimization using an Nvidia GeForce RTX 2080 TI Graphics Card on CIFAR100. } 
\label{tab:rl_running_time}
\vskip 0.15in
\begin{center}
\begin{small}
\begin{sc}
\begin{tabular}{l|c|c}
\toprule
  & Running Time (s) & Accuracy (\%)\\
\midrule
RAR ($K=10$) & 1294 $\pm$ 255& 27.3  $\pm$ 0.3 \\
RAR ($K=20$)&3499  $\pm$ 406& 27.3  $\pm$ 0.5  \\
RL-RAR ($K=19.8$) & 6137 $\pm$ 34 & 29.2 $\pm$ 0.3  \\
\bottomrule
\end{tabular}
\end{sc}
\end{small}
\end{center}
\vskip -0.1in
\end{table}

\subsection{Offline Continual Learning}
\label{sec:offline_cl}
{\color{black}Although this paper is mostly focused on online continual learning, some of the analysis and discussion are also of independent interest to offline continual learning. Specifically, from a theoretical perspective, the empirical risk minimization of offline rehearsal is shown in Proposition 2 in Section~\ref{sec:theory} and its augmented risk is shown in Proposition 3. Two conclusions can be drawn from this analysis. First, the risk of memory overfitting in offline rehearsal is also related to the problem characteristic, the ratio $\lambda$ between task size and memory size. Second, augmentation can help with offline rehearsal since the orbit-averaging operation in the augmented empirical risk can reduce both the model variance and generalization error. From an empirical perspective, Table~\ref{tab:offline_cl} shows the performance of offline ER with and without augmentation in four datasets. We use 50 epochs and a memory size of 2000. For all four datasets, offline ER with augmentation achieves a significant performance gain over offline ER without augmentation. More interestingly, compared to datasets with a small $\lambda$ (e.g., CLRS with $\lambda=1.125$) we observe that the datasets with a higher task-to-memory size ratio (e.g., CORE50 with $\lambda=6$) tend to benefit more from augmentation, due to the increased risk of memory overfitting.}
\begin{table}[t]
\caption{\color{black}Performance of offline rehearsal with and without augmentation. } 
\label{tab:offline_cl}
\begin{center}
\begin{small}
\begin{sc}
\resizebox{\textwidth}{!}{%
\begin{tabular}{l|cccc}
\toprule
 & Seq-CIFAR100 & Seq-Mini-ImageNet  & CORE50-NC & CLRS25-NC\\
\midrule
Offline ER w/o Aug  
& 17.0 $\pm$ 0.6 
& 20.4 $\pm$ 0.6  
& 30.2 $\pm$ 1.4
& 33.5 $\pm$ 1.6 \\
Offline ER w/ Aug   
& 28.3 $\pm$ 0.6 
& 32.6  $\pm$ 0.1
& 44.5 $\pm$ 1.3
& 35.7 $\pm$ 0.8 \\
Gains 
& 11.3 $\uparrow$
& 12.2 $\uparrow$
& 14.3 $\uparrow$
& 2.2 $\uparrow$ \\
\bottomrule
\end{tabular}
}
\end{sc}
\end{small}
\end{center}
\end{table}
\end{document}